%% file: acl2023.tex
\pdfoutput=1
\documentclass[11pt]{article}
\usepackage{microtype}
\usepackage{graphicx}
\usepackage{subfig}
\usepackage{booktabs} % for professional tables
\usepackage{multirow}

\usepackage[]{ACL2023}

\usepackage{amsmath}
\usepackage{amssymb}
\usepackage{mathtools}
\usepackage{amsthm}
\usepackage{pifont}
\usepackage{xspace}
\usepackage{enumitem}

\usepackage[capitalize,noabbrev]{cleveref}

\theoremstyle{plain}
\newtheorem{theorem}{Theorem}[section]

\theoremstyle{definition}
\newtheorem{definition}[theorem]{Definition}

\theoremstyle{remark}

\usepackage[textsize=tiny]{todonotes}

\DeclareMathOperator*{\argmax}{arg\,max}

\usepackage{times}
\usepackage{latexsym}

\usepackage[T1]{fontenc}
\usepackage[utf8]{inputenc}

\usepackage{microtype}

\usepackage{inconsolata}

\newcommand{\cf}{{\it cf.}\xspace}
\newcommand{\eg}{{\it e.g.}\xspace}

\newcommand{\ie}{{\it i.e.}\xspace}
\title{Machine Unlearning of Pre-trained Large Language Models}

\author{Jin Yao$^1$ \ \ \ Eli Chien$^2$ \ \ \ Minxin Du$^3$\thanks{\; Work partially done when at CUHK.} \ \ \ Xinyao Niu$^4$ \ \ \ Tianhao Wang$^1$ \\
\ \ \ \textbf{Zezhou Cheng}$^1$ \ \ \ \textbf{Xiang Yue}$^{5}\thanks{\; Corresponding Author}$\\[0.5em]
$^1$University of Virginia \ \ \ $^2$Georgia Institute of Technology \\$^3$The Hong Kong Polytechnic University \ \ \ $^4$University of Melbourne \\  $^5$Carnegie Mellon University \\ 
\small
\texttt{rry4fg@virginia.edu} \ \ \ 
\texttt{ichien6@gatech.edu} \ \ \ 
\texttt{xyue2@andrew.cmu.edu}
}

\begin{document}
\maketitle
\begin{abstract}
 This study investigates the concept of the \textit{`right to be forgotten'} within the context of large language models (LLMs). 
 % exemplified by The New York Times’ legal action against OpenAI and Microsoft for purported copyright violations in LLMs training datasets. 
 We explore machine unlearning as a pivotal solution, with a focus on pre-trained models--a notably under-researched area. Our research delineates a comprehensive framework for machine unlearning in pre-trained LLMs, encompassing a critical analysis of seven diverse unlearning methods. 
 Through rigorous evaluation using curated datasets from arXiv, books, and GitHub, we establish a robust benchmark for unlearning performance, demonstrating that these methods are over $10^5$ times more computationally efficient than retraining.
 % We rigorously evaluate these methods against curated datasets from arXiv, books, and GitHub codes, providing a robust benchmark for unlearning performance. Our findings show these unlearning methods surpass retraining in computational efficiency by over $10^5$ times. 
 Our results show that integrating gradient ascent with gradient descent on in-distribution data improves hyperparameter robustness. We also provide detailed guidelines for efficient hyperparameter tuning in the unlearning process.
 Our findings advance the discourse on ethical AI practices, offering substantive insights into the mechanics of machine unlearning for pre-trained LLMs and underscoring the potential for responsible AI development.\footnote{Our dataset and code is available at 
 % \url{https://anonymous.4open.science/r/Unlearning_LLM-503F}}
 \url{https://github.com/yaojin17/Unlearning_LLM}}
\end{abstract}

\input{1-intro}

\input{2-formulation}

\input{3-methods}
\input{4-experiments}
\input{5-related}
\input{6-conclusion}

\section*{Limitations}
This work primarily focuses on conducting experiments with the Yi-6B model. 
A significant challenge arises since most LLMs do not open-source their pre-training data, making the collection of forget sets infeasible. 
We encourage future research to investigate the applicability of unlearning processes to other models, including those of larger sizes such as 13B or 70B, or more complicated architecture such as the mixture of experts. Additionally, our experiments are mainly conducted on three specific pre-training data domains. Future research should aim to explore unlearning across other domains, including Wikipedia and News. 

Moreover, our work concentrates on unlearning copyrighted content from LLMs. Future studies could expand our methodologies to address other challenges, such as unlearning biases or harmful outputs in LLMs. 
Given that our methods are non-convergent and may reduce model utility until convergence, the adjustment of hyperparameters becomes crucial for ideal unlearning results. While our guidelines simplify and streamline this process, we hope that future research will develop convergent methods that are less dependent on hyperparameter adjustments.

This study emphasizes the practical aspects of approximate unlearning in LLMs. However, since distinguishing between member and non-member data in LLMs remains challenging, the evaluation of approximate unlearning can be complex. A more principled theoretical investigation of unlearning in LLMs is necessary for future work. Additionally, more powerful MIA methods and alternative methods beyond MIA are needed for evaluating unlearning methods in the context of LLMs. While our focus is on unlearning data within a single domain while maintaining the unlearned model's general capability, exploring the unlearning of task-specific data and assessing the impact on the unlearned model's performance in that task could be an interesting and important direction for future research.

\section*{Ethics Statement}
In this work, we focus on unlearning pre-trained generative LLMs. Our goal is to enable LLMs to selectively forget particular training sequences while preserving the model's utility. This approach aims to address ethical concerns, including copyright infringement and privacy breaches. The evaluation datasets are compiled from publicly accessible sources, adhering to the licenses associated with the collected data. We also encourage researchers and developers to use our methods responsibly and ethically.  

\section*{Acknowledgements}
The authors thank the Yi model and its developer for providing a sample of pre-training data, along with permission for its use and open-sourcing.
Wang was partially supported by NSF OAC-2319988.

\bibliographystyle{acl_natbib}
\bibliography{custom}

\clearpage
\appendix
\input{app-ablation}
\input{app-random_seed}

\input{app-related}

\end{document}

%% file: 1-intro.tex
\section{Introduction}
\begin{figure*}[!t]
  \centering
  \includegraphics[width=1\linewidth]{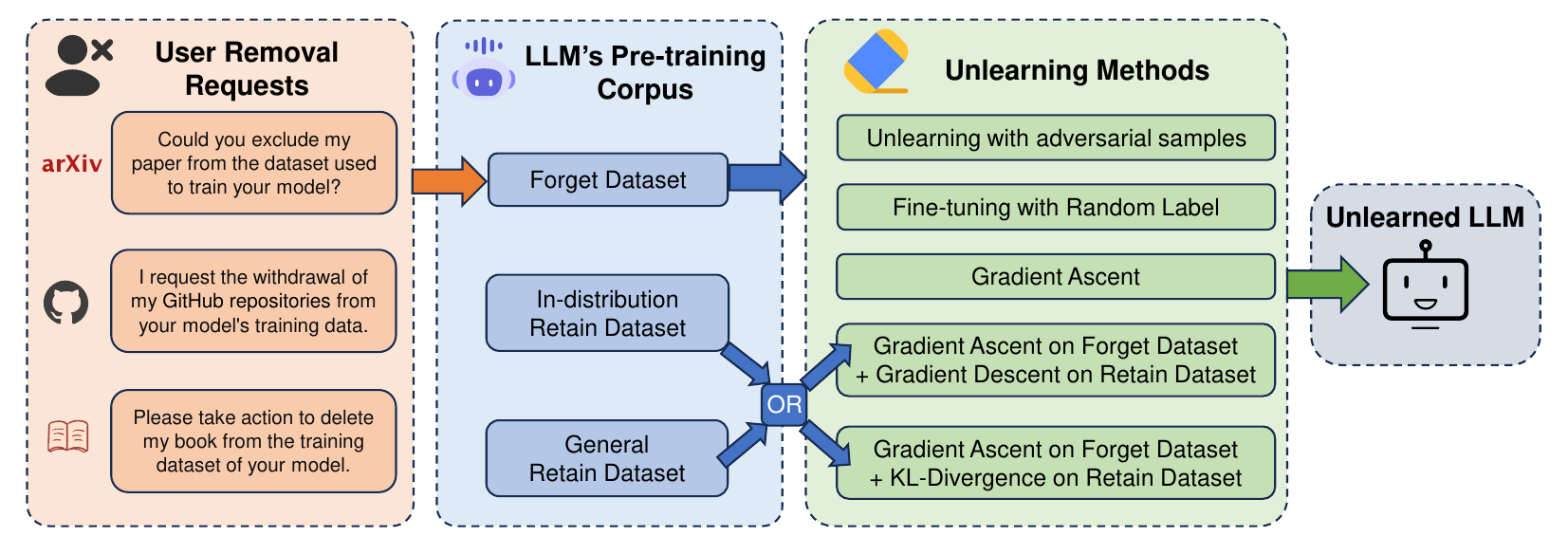}
  % \vspace{-20mm}
  \caption{Overview of unlearning pre-trained LLMs to address user removal requests. }
  \label{fig:overview}
\end{figure*}

Large language models (LLMs) have seen remarkable advancements, attributable to training on extensive and diverse datasets~\cite{nips/Ouyang0JAWMZASR22,wei2022emergent, corr/TouvronMSA+23,wu2023bloomberggpt,liang2023helm}. Yet, the reliance on~massive data pools has raised significant ethical concerns, particularly when such data include sensitive, private, or copyrighted material~\cite{li2022large,shi2023detecting,li2023meticulously,yang2024code}. 

A prominent example of these issues is the recent lawsuit filed by The New York Times\footnote{\url{https://nytco-assets.nytimes.com/2023/12/NYT\_Complaint\_Dec2023.pdf}} against OpenAI.
%and Microsoft. 
The lawsuit, responding to the alleged use of millions of articles from The Times in training LLMs like ChatGPT, highlights the critical issue of copyright infringement in LLMs' development.

In response to these ethical challenges, the concept of \emph{machine unlearning} has emerged as a potential remedy. 
It entails systematically removing specific data from a model’s training, ensuring its operation as though the data had never been included~\cite{sp/BourtouleCCJTZL21}. This approach mitigates the ethical issues stemming from the pre-trained data in LLMs, aligning the technology with evolving legal and ethical standards.

% Despite its potential, current research on machine unlearning in the realm of LLMs has been predominantly confined to the fine-tuned model~\cite{corr/KumarGR22, corr/ChenY23, maini2024tofu}. This focus has limitations, as fine-tuning models on a retained dataset is often feasible, rendering fine-tuning phase unlearning less critical. 
Despite its potential, current research on machine unlearning in the realm of LLMs has been predominantly confined to the fine-tuned model~\cite{corr/KumarGR22, corr/ChenY23, maini2024tofu}. 
While fine-tuning datasets typically contain token numbers on the magnitude of $10^{6}$ or less, pre-training corpora are on the magnitude of $10^{12}$.
Consequently, retraining on the retained set of a fine-tuning dataset based on a pre-trained model is often feasible, whereas retraining the pre-trained model from scratch is impractical. Therefore, this focus on unlearning fine-tuned models has limitations and is less critical. 
The real challenge, and our focus in this paper, is the \textit{unlearning of pre-trained LLMs}. 
This challenge is compounded by several factors: 
1) the need to adapt existing unlearning methods from other fields to pre-trained LLMs, 
2) the general lack of public availability of pre-trained data used to develop LLMs, 
and 3) the absence of directly comparable baselines due to the exorbitant costs of retraining pre-trained LLMs.

Our paper addresses them through several key contributions. 
We first define the problem of machine unlearning for pre-trained LLMs and propose a unified formulation consolidating prior arts under a single unlearning objective. 
We then investigate seven different unlearning methods in the context of LLMs. 
To benchmark the unlearning performance, we compile three datasets from sources, including arXiv, books, and GitHub code. 
Recognizing the impracticality of retraining pre-trained models, we propose an approximate retraining method using an in-distribution, unseen dataset to simulate the performance of a retraining baseline. 

Besides, previous studies on machine unlearning in LLMs have been limited to small-scale experiments, unlearning at most 128 samples, single corpus sources, or short context lengths~\cite{acl/JangYYCLLS23, corr/EldanR23, yao2023large}. 
In contrast, our work unlearns thousands of chunks, each 4096 tokens in length, from a diverse range of sources across three domains, presenting a more realistic and challenging scenario. 
Fig.~\ref{fig:overview} illustrates a process overview for unlearning pre-trained LLMs to address user removal requests. Our main contributions and findings are summarized below. 
\begin{itemize}[leftmargin=*]
    \item We structure a unified unlearning framework for LLMs, from which seven unlearning methodologies are derived and adapted to LLMs.
    \item We introduce an approximate retraining evaluation baseline to bypass the impracticality of retraining LLMs. Experiments on three domains demonstrate the efficacy of our methods. 
    \item We open-source our dataset, including samples from a real pre-training dataset, which is typically a private and closed-source asset. 
    \item  Gradient ascent combined with gradient descent on in-distribution data shows greater hyperparameter robustness. We offer guidelines to efficiently fine-tune hyperparameters for other methods to streamline and make unlearning more feasible. 
    % critical to unlearning LLMs, aiming to streamline and make unlearning more feasible. 
\end{itemize}

We aim to offer a comprehensive solution to unlearning in pre-trained LLMs, contributing to developing more ethical and responsible AI systems.

%% file: 2-formulation.tex
\section{Problem Formulations}
\label{sec:formulation}

Let $\mathcal{D} = \{x_i\}_{i=1}^N$ be a training corpus containing $N$ sequences, where $x_{i \in [N]} $ is a sequence of $t_i$ tokens \( w_1^i, w_2^i, ..., w_{t_i}^i \).
With a slight abuse of notation, we use $M$ to denote both the model itself and its weights.
This work focuses on generative LLMs $M$ that are typically trained using the next-token prediction, characterizing the conditional probability given prompts: $P_{M}(w_{t+1} | w_1, w_2, ..., w_t)$.
We denote $A$ as a randomly initialized training algorithm $M \leftarrow A(\mathcal{D})$, where the training objective is to minimize the negative log-likelihood: 
$$\mathcal{L}(P_{M};\mathcal{D}) = -\sum_{x_i \in \mathcal{D}} \sum_{t=1}^{t_i} \log P_{M}(w_{t+1}^i | w_1^i, ..., w_t^i).$$

We call the model designated for unlearning the \emph{vanilla} model. 
We denote a forget set of sequences to be unlearned as $\mathcal{U} \subset \mathcal{D}$.
To remove the effect of $\mathcal{U}$, we consider an unlearning algorithm $\widehat{A}$ that applies to $\mathcal{A}(\mathcal{D})$ and outputs an unlearned model~$M^\prime$.

Motivated by differential privacy~\cite{tcc/DworkMNS06}, \citet{nips/GinartGVZ19} formulated the probabilistic notion of unlearning using $(\epsilon, \delta)$-closeness of   distributions~\cite{icml/GuoGHM20, nips/SekhariAKS21}.
Informally, it requires that the output distributions of $\widehat{A}$ and $A$ run over $(\mathcal{D} \setminus \mathcal{U})$ to be similar.
We call it \emph{exact} unlearning if the distributions are identical (\ie, $\epsilon = \delta = 0$); \emph{approximate} unlearning, otherwise.
A na\"ive solution for exact unlearning is just retraining on $\mathcal{D} \setminus \mathcal{U}$ for each $\mathcal{U}$ from scratch.
However, it is prohibitively expensive for LLMs, incurring gigantic computation costs and carbon footprints~\cite{corr/LuccioniVL22,corr/ZhangFTP+23}.

% A na\"ive solution for $\widehat{A}$ is to completely retrain a model from scratch on $\mathcal{D} \setminus \mathcal{U}$ for each deletion request $\mathcal{U}$.
% %without considering the previous model $A(\mathcal{D})$. 
% This is also known as \emph{exact} unlearning. However, it is impractical for LLMs due to high computational costs and carbon footprint induced by re-training~\cite{corr/LuccioniVL22}.

Deriving theoretical guarantees for LLMs is also non-trivial, as the underlying transformer architecture is not convex or Lipschitz~\cite{icml/KimPM21}.
Pragmatically, an active line of research~\cite{cvpr/GolatkarAS20,corr/ChenY23,corr/KurmanjiTT23,nips/JiaLRYLLSL23} only requires the empirical performance (\eg, classification accuracy) of retrained and unlearned models to be similar.
In our context, we can resort to \emph{perplexity} and ensure
$$\mathbb{E}P_{M^\star} \approx \mathbb{E}P_{M^\prime} \ \text{with} \  M^\star \leftarrow A(\mathcal{D} \setminus \mathcal{U}),$$
where $M^\star$ represents the model trained on $\mathcal{D}\setminus \mathcal{U}$.
Besides, we require that the validation performance of $M^\prime$ on $\mathcal{D} \setminus \mathcal{U}$ and $\mathcal{U}$ is similar to that of the vanilla model on $\mathcal{D} \setminus \mathcal{U}$ and unseen data, respectively.

%% file: 3-methods.tex
\section{Unlearning Methods}
\label{sect:methods}
\subsection{Overview}
% \du{i) Seems better to introduce Eq.~(1) in Section 2 as a ``counterpart'' of the training objective (as we will remove Def. 1 and other ``theoretical'' parts)?
% ii) Outline the methods (corresponding to those in experiments) and their relationship given in later subsections.
% iii) Remove ``first order methods'' and use 3 subsections corresponding to a) Fine-tuneing, b) Gradient ascent + descent, c) Gradient ascent + KL divergence?
% iv) introduce KL divergence somewhere, e.g., preliminaries?}
As discussed in Section~\ref{sec:formulation}, the objective of LLM unlearning is to \emph{ensure that the model effectively forgets designated token sequences while still preserving its performance on the retain set}. 
To achieve this goal, we propose an approximate unlearning framework for LLMs using next-token prediction. 
To unlearn sequences in $\mathcal{U}$, we update the current model $M$ using the gradient derived from 
\begin{align}\label{eq:general_fou}
    & \sum_{w\in \mathcal{U}}\sum_{t=1}^{T}\mathbb{E}_{q_t\sim Q_{w_t}} \log P_{M}(q_{t} | w_1, w_2, ..., w_{t-1}) \nonumber\\
    & + \sum_{z\in \mathcal{R}}\sum_{t=1}^T\log P_{M}(z_t | z_1, z_2, ..., z_{t-1}),
\end{align}
where $\mathcal{R}\subseteq \mathcal{D}\setminus \mathcal{U}$ 
%is some subset of retain sequences 
and $Q_{w_t}$ is a set of distributions over token universe $\mathcal{W}$ depending on $w_t$, which we call \emph{reference distributions}. 
While Eq.~\eqref{eq:general_fou} appears complicated, we demonstrate that several iconic unlearning methods are its instances in Section~\ref{subsect:unlearning methods}.

Most existing unlearning approaches~\cite{cvpr/GolatkarAS20,corr/ChenY23,corr/KurmanjiTT23,nips/JiaLRYLLSL23} target at (image) classification scenarios.
Nevertheless, some of them can be adapted to unlearning token sequences used to train (generative) LLMs with slight modifications.  

Below, we focus on those \emph{first-order approximate} unlearning methods that only exploit gradient information and are often more efficient than exact unlearning and second-order designs~\footnote{We present a detailed review on exact unlearning and second-order unlearning methods in Appendix~\ref{app:related_work}}.
Their general formulation is given in Eq.~\eqref{eq:general_fou}, which can be extended for new unlearning methods.
We will discuss how to specialize it to each method revisited below and the corresponding pros and cons.

Notably, some methods either consider forgetting an entire class~\cite{tarun2021classunlearn} or need to store all intermediate model/gradient information during training~\cite{sp/BourtouleCCJTZL21}.
However, the former is not directly applicable to LLMs, and the latter is too expensive regarding memory.
We exclude them from our discussion. 

% They are not directly applicable to LLMs, and thus we exclude them from our discussion. 
% On the other hand, some unlearning methods require storing all intermediate model/gradient information during training~\cite{xxx}. 
% These approaches are too expensive in terms of memory for LLM unlearning and thus we ignore them in our manuscript.

\subsection{Approximate Unlearning Methods}
\label{subsect:unlearning methods}
\subsubsection{Gradient Ascent (or Negative Gradient)} 
Derived from the general framework outlined in Eq.~\eqref{eq:general_fou}, if we ignore the second term, set $Q_{w_t} = \delta_{w_t}$, and multiply $-1$ to the gradient, we arrive at the unlearning strategy known as gradient ascent or negative gradient~\cite{cvpr/GolatkarAS20,nips/JiaLRYLLSL23,acl/JangYYCLLS23}.
Here, $\delta_{w_t}$ is the delta function at $w_t$ such that $q_t\sim Q_{w_t}$ means $q_t = w_t$ with probability $1$. 
The intuition is that $M$ has been trained with $\mathcal{U}$, while the retrained model $M_r^{\mathcal{U}}$ never sees $\mathcal{U}$. Thus, the loss of $M$ on $\mathcal{U}$ is lower than that of $M_r^{\mathcal{U}}$, but the loss should be similar if $|\mathcal{U}|$ is limited.
Unfortunately, it is known from the literature that if we perform gradient ascent for too many epochs, the model $M$ will also potentially forget the information about $\mathcal{D}\setminus \mathcal{U}$, thus leading to poor utility. 
In practice, researchers often only apply gradient ascent in a few epochs.

\subsubsection{Fine-tuning with Random Labels} 
Alternatively, if we ignore the second term of Eq.~\eqref{eq:general_fou}, and set $Q_{w_t}$ to be a uniform distribution over all possible token sets $\mathcal{W}$, we arrive at the strategy known as fine-tuning with random labels, as proposed by~\citet{cvpr/GolatkarAS20} for classification problems. 
The intuition for this strategy is that a model not seeing $\mathcal{U}$ should act as random guessing. 
While it may seem reasonable at first glance, we argue that uniform distribution for $Q_{w_t}$ is not universally appropriate. 
For instance, consider the case of two duplicated sequences: one to be unlearned and the other to be retained.
Apparently, the retrained model should not act as random guessing on this sequence.  
In practice, convergence on random labels often leads to a marked decrease in both utility and performance, limiting this method to a brief period of weight adjustment, akin to the earlier mentioned gradient ascent method.
% In practical applications, training a model to converge with random labels invariably results in significantly diminished utility and performance.
% Consequently, this method is generally restricted to a few epochs of weight updating, similar to the above gradient ascent.

% \textbf{Fine-tune with incompetent teacher~\cite{aaai/ChundawatTMK23,corr/ZhangWCSZX23}:} 
~\citet{aaai/ChundawatTMK23,corr/ZhangWCSZX23} propose to set $Q_{w_t}=P_{M_{\text{rand}}}(w_t|w_1, ..., w_{t-1})$, where $M_{\text{rand}}$ is a randomly initialized model (known as incompetent teacher). 
Their intuition is similar to fine-tuning with random labels, where $M_{\text{rand}}$ does not contain information about $\mathcal{U}$. 
Essentially, these two methods are equivalent, but fine-tuning with random labels is more direct and efficient, so we only adapt it to LLMs. 

% Nevertheless, it will suffer from a similar issue as fine-tuning with random labels. \jin{Is this method the same as Fine-tune with random label?}

\subsubsection{Unlearning with Adversarial Samples} 
This approach is originally proposed for classification tasks~\cite{corr/ChaCHLML23}. 
We adapt it to our context below. 
For simplicity, let us assume only one sequence $w_1,\ldots,w_T$ to be unlearned. 
We generate adversarial samples $\{a_t\}$ for each $t$ such that they are close to $w_t$ but can confuse $M$ the most
\begin{align}\label{eq:adv}
    & a_t= \argmax_{a\neq w_t} P_{M}(a|w_1, w_2,\ldots, w_{t-1}).
\end{align}
Originally,~\citet{corr/ChaCHLML23} proposes choosing adversarial samples $a$ within a small radius to $w_t$ in some metric space for classification tasks. 
Yet, it is non-trivial to adapt this strategy to LLMs. We thus choose a most likely token $a$ other than $w_t$. 

% Basically, we want to search for a token $a$ (other than $w_t$) closest to the ``decision boundary'' defined by $M$. 
% For classification problems, ~\citet{corr/ChaCHLML23} also propose choosing adversarial samples $a$ within a small radius to $w_t$ in some metric space. 
% Yet, it is non-trivial to adapt this strategy to LLMs. 
%So, we decide to ignore this step for simplicity. 

To unlearn all training sequences in $\mathcal{U}$, we fine-tune $M$ using Eq.~\eqref{eq:general_fou} while ignoring the second term in Eq.~\eqref{eq:general_fou} and choosing $Q_{w_t}=\delta_{a_t}$ and $a_t=\argmax_{a\neq w_t} P_{M}(a|w_1, w_2,\ldots, w_{t-1})$. 
%There are a few practical details. 
The raw approach~\cite{corr/ChaCHLML23} uses $K-1$ adversarial samples for each training sample to be unlearned in $K$-class classification problems.
Directly using it is not suitable, as generating $|\mathcal{W}|-1$ adversarial samples for unlearning one token is impractical in our setting. 
Hence, we simplify it via $Q_{w_t}=\delta_{a_t}$ and $a_t=\argmax_{a\neq w_t} P_{M}(a|w_1, w_2,\ldots, w_{t-1})$. 
Other choices in the same spirit, such as replacing $\argmax_{a\neq w_t}$ with $\text{top-k}_{a\neq w_t}$, are also feasible.

\subsubsection{Gradient Ascent + Descent or KL Divergence on Retained Set} 
On the other hand, 
% if we ignore the first term in Eq.~\eqref{eq:general_fou}, we arrive at the unlearning method called fine-tuning on the retained set. 
disregarding the first term in Eq.\eqref{eq:general_fou} leads to a strategy known as fine-tuning on the retained set. 
By updating $M$ with this strategy on $\mathcal{D}\setminus \mathcal{U}$ until convergence, we can achieve the effect of retraining from scratch. 
For efficiency, researchers typically fine-tune $M$ on a small subset $\mathcal{R}\subseteq \mathcal{D}\setminus \mathcal{U}$  over a few epochs. 
However, this becomes impractical for LLMs due to the large volume of pre-training data relative to the data designated for unlearning. 
% Although this approach is not used independently in this work, it will be combined with gradient ascent techniques. Such a hybrid approach effectively balances unlearning effectiveness and overall utility by optimizing both terms in Eq.~\eqref{eq:general_fou}. 
While not utilized independently here, this method is integrated with gradient ascent techniques, forming a hybrid approach that optimizes both terms in Eq.\eqref{eq:general_fou} to balance unlearning effectiveness with utility. 
To optimize the second term, we adopt both direct gradient ascent and KL-divergence constraint methods, outlined in the prior work~\cite{yao2023large}. Moreover, we assess the impact of different data types for the second term, including general pre-training data and domain-specific data matching the unlearning set, termed in-distribution data.

\begin{table*}[!t]
\resizebox{\linewidth}{!}{%
\begin{tabular}{@{}l|ccc|cc|ccccc@{}}
\toprule
\multirow{2}{*}{Models} & \multicolumn{3}{c|}{Forget Set} & \multicolumn{2}{c|}{Retain Set} & \multicolumn{5}{c}{Downstream Task Accuracy $\uparrow$} \\ \cmidrule(l){2-11} 
 & ACC$\downarrow$ & PPL$\uparrow$ & MIA$\downarrow$ & ACC$\uparrow$ & PPL$\downarrow$ & MMLU & ARC & HumanEval &  GSM8K&Avg. \\ \midrule
 Vanilla Model & 69.02 & 3.65 & 50.77 & 52.68 & 9.24 & 63.37 & 68.49 & 16.46 & 33.59 & 45.48\\
Approximate Retrain & 68.98 & 3.69 & - & - & - & - & - & - & - & -  \\
Gradient Ascent &68.79  &3.70  & 50.28 &52.66  & 9.26 &63.45 &68.77   &15.85 &34.04 &  45.53 \\
Fine-tuning with Random Labels & 68.92 & 3.69 & 50.55 & 52.67 &9.25  & 63.37 & 68.38 &14.02 &32.22 & 44.50  \\
Unlearning with Adversarial Samples &68.87 &3.69 & 50.52 &52.68 & 9.25 & 63.32 & 68.74 &15.24 & 33.13& 45.11 \\ \midrule
Gradient Ascent + Descent on retain set &  &  &  &  &  &  &  &  &  &  \\
- Descent on in-distribution data &68.87  & 3.69 & 50.18 &52.66 & 9.26 &63.32  &68.52  & 15.24 &33.74&45.21  \\
- Descent on general data &68.81 & 3.69 & 50.33 &52.93 &9.04  &63.40  &67.87  & 15.24 &33.13  & 44.91 \\ \midrule

Gradient Ascent + KL divergence &  &  &  &  &  &  &  &  &  &  \\
- KL on in-distribution data &68.82  &3.69  & 50.29 &52.65 &9.27 &63.40  &68.57&15.24 &33.89 &45.28 \\
- KL on general data &68.79 & 3.70  & 50.25 & 52.65 & 9.27 & 63.27 & 68.38 & 15.85 & 33.81 & 45.33 \\
\bottomrule
\end{tabular}%
}
\caption{Overall results of unlearning an open-source LLM on a subset of pre-training data (500 \textbf{arXiv papers})}
\label{tab:unlearn_train_data_results_arxiv}
\end{table*}

\begin{table*}[!t]
\resizebox{\linewidth}{!}{%
\begin{tabular}{@{}l|ccc|cc|ccccc@{}}
\toprule
\multirow{2}{*}{Models} & \multicolumn{3}{c|}{Forget Set} & \multicolumn{2}{c|}{Retain Set} & \multicolumn{5}{c}{Downstream Task Accuracy $\uparrow$} \\ \cmidrule(l){2-11} 
 & ACC$\downarrow$ & PPL$\uparrow$ & MIA$\downarrow$ & ACC$\uparrow$ & PPL$\downarrow$ & MMLU & ARC & HumanEval &  GSM8K & Avg. \\ \midrule
 Vanilla Model & 80.65 & 2.40 & 81.93 & 52.68 & 9.24 & 63.37 & 68.49 & 16.46 & 33.59 & 45.48 \\
Approximate Retrain & 72.91 & 3.42  & - & - & - & - & - & - & - & -  \\
Gradient Ascent & 78.19 & 3.53  &74.28  & 52.60 & 9.31 & 63.45 &68.40  & 14.63 & 35.10 & 45.40 \\
Fine-tuning with Random Labels &78.00  & 3.12 &80.55  & 52.50 & 9.47 & 62.45 & 67.02 & 10.98 & 29.49 & 42.48  \\
Unlearning with Adversarial Samples & 75.09& 3.40 & 79.51 & 52.54 & 9.41 &  62.36 & 67.33 & 9.76 & 31.39 &  42.71  \\ \midrule
Gradient Ascent + Descent on retain set &  &  &  &  &  &  &  &  &  &  \\
- Descent on in-distribution data & 76.88 & 3.45 & 76.75 & 52.48 & 9.38 & 62.31 & 66.77 & 2.44 & 31.01 &  40.63 \\
- Descent on general data & 78.79 & 3.57  &75.61  & 53.03 & 9.00  & 63.15 & 67.62 & 14.63 & 33.51 & 44.73 \\ \midrule

Gradient Ascent + KL divergence &  &  &  &  &  &  &  &  &  &  \\
- KL on in-distribution data &78.78 &3.51  &76.19  & 52.61 & 9.31 & 63.40 & 68.21 & 14.63 &34.95  & 45.30 \\
- KL on general data & 78.68& 3.58 & 75.42 & 52.60 & 9.31  & 63.32 & 68.07 & 14.02 & 34.72 & 45.03 \\
\bottomrule
\end{tabular}%
}
\caption{Overall results of unlearning an open-source LLM on a subset of pre-training data (2K \textbf{GitHub code repository files}). Results under multiple training random seeds are displayed in Table~\ref{tab:unlearn_train_data_results_github_random_seed}.}
\label{tab:unlearn_train_data_results_github}
\end{table*}

\begin{table*}[!t]
\resizebox{\linewidth}{!}{%
\begin{tabular}{@{}l|ccc|cc|ccccc@{}}
\toprule
\multirow{2}{*}{Models} & \multicolumn{3}{c|}{Forget Set} & \multicolumn{2}{c|}{Retain Set} & \multicolumn{5}{c}{Downstream Task Accuracy $\uparrow$} \\ \cmidrule(l){2-11} 
 & ACC$\downarrow$ & PPL$\uparrow$ & MIA$\downarrow$ & ACC$\uparrow$ & PPL$\downarrow$ & MMLU & ARC & HumanEval &  GSM8K& Avg. \\ \midrule
 Vanilla Model &55.26 & 7.62   &74.03  & 52.68 & 9.24 & 63.37 & 68.49 & 16.46 & 33.59& 45.48 \\
Approximate Retrain & 50.65  & 10.11   & - & - & - & - & - & - & - & - \\
Gradient Ascent & 52.47 & 9.64   &58.47  & 52.45  & 9.40  & 63.32 & 68.66 & 16.46 &32.90  & 44.91 \\
Fine-tuning with Random Labels & 51.9  &  10.19    &63.69  & 52.56  &  9.39 & 63.05 &68.01  & 16.46 & 29.64 & 44.29 \\
Unlearning with Adversarial Samples & 52.07 &  10.02   & 63.60 &  52.59 &  9.35 & 63.08 & 68.18 & 16.46 & 31.39 & 44.78 \\ \midrule
Gradient Ascent + Descent on retain set &  &  &  &  &  &  &  &  &  &  \\
- Descent on in-distribution data & 50.07  & 10.27   &56.39  & 52.34  &  9.41  & 63.08 &67.70  & 17.68 & 29.80 &44.57  \\
- Descent on general data & 52.49 & 10.35  & 69.81 & 52.88 & 9.06 & 63.33 & 67.78 & 16.46 & 32.83 & 45.10 \\ \midrule

Gradient Ascent + KL divergence &  &  &  &  &  &  &  &  &  &  \\
- KL on in-distribution data & 52.42 &  10.02   & 64.02 & 52.52  & 9.35  &63.50  & 68.80& 16.46 &33.59  &45.59  \\
- KL on general data &52.85  & 9.71  &62.61  & 52.58  & 9.31  & 63.32 & 68.55 & 15.24 & 32.98 & 45.02 \\
\bottomrule
\end{tabular}%
}
\caption{Overall results of unlearning an open-source LLM on a subset of pre-training data (100  \textbf{Books})}
\label{tab:unlearn_train_data_results_books}
\end{table*}

% \begin{table*}[!t]
% \resizebox{\linewidth}{!}{%
% \begin{tabular}{@{}l|cc|cc|ccccc@{}}
% \toprule
% \multirow{2}{*}{Models} & \multicolumn{2}{c|}{Forget Set} & \multicolumn{2}{c|}{Generalization Set} & \multicolumn{5}{c}{Downstream Task Accuracy$\uparrow$} \\ \cmidrule(l){2-10} 
%  & Percentage$\downarrow$ & PPL$\uparrow$ & Percentage$\downarrow$ & PPL$\uparrow$ & NQ & MMLU & BoolQ & PIQA & HumanEval \\ \midrule
% Vicuna-13B v1.5  & 50.1 & 50.1 & 50.1 & 50.1 & 50.1 & 50.1 & 50.1 & 50.1 & 50.1 \\
% Gradient Ascend &  &  &  &  &  &  &  &  &  \\
% Fine-tune with Random Label &  &  &  &  &  &  &  &  &  \\
% Fine-tuned with Reference Model &  &  &  &  &  &  &  &  &  \\ \bottomrule
% \end{tabular}%
% }
% \caption{Overall results of unlearning jailbreaking generations (52K flagger conversations from LMSYS-Chat1M) using different methods}
% \label{tab:unlearn_harmful_generations}
% \end{table*}

%% file: 4-experiments.tex
\section{Experiments}
% We select one representative application scenario for training data unlearning and behavioral unlearning respectively. 

% \subsection{Type I: Training Data Unlearning} 
\subsection{Background}
 Here, we select removing copyrighted data from the pre-trained model as a representative scenario. 
 LLMs have the potential to internalize and reproduce copyrighted content unintentionally. 
 It~poses legal challenges and ethical dilemmas, especially when the model's outputs mimic or rephrase the protected material. 
 When it comes to light that copyrighted data has been assimilated into an LLM's training set, machine unlearning techniques can be mobilized to facilitate the model's ``forgetfulness'' regarding this specific content. 
 Hence,~the model's subsequent outputs are safeguarded against undue influences of the copyrighted material, fostering a more compliant and ethical use of data.
% \paragraph{Dataset.} We consider unlearning a branch of arXiv papers from the Yi-6B\footnote{https://huggingface.co/01-ai/Yi-6B} large language models. Although arXiv provides open access to preprints, in many cases the copyright of each individual preprint remains with its authors or the rights holders. Imagine a case where the authors of the paper would like to erase their preprints from the pre-trained LLMs (``the right to be forgotten''). To mimic this process, we sample 100K arXiv articles as the \textit{forget set} from Redpajama \cite{together2023redpajama}, an open-source pre-training data collection to reproduce Llama.

\subsection{Evaluation Metrics}
 We focus on evaluating the unlearned models from 

 \noindent 1) \textbf{Performance on the Forget Set}: The model should not be able to predict correctly on the forget set, or its performance should degrade to the same level as the test set.

 \noindent 2) \textbf{Performance on the Retain Set}: Ideally, the model's performance on this set should not degrade significantly, indicating that the unlearning process did not adversely affect the data it should remember. 
 The performance assessment is conducted by measuring the model's accuracy and perplexity on both the forget and retain sets. 
 
 \noindent 3) \textbf{Performance on General Downstream Tasks:} We can evaluate the performance on some general downstream tasks, which can provide insights into the model's overall capability post-unlearning. 
 The model's performance on these tasks is expected not to downgrade too much compared with the model before unlearning. 
 The downstream tasks considered include Massive Multitask Language Understanding (MMLU)~\cite{mmlu}, the ARC Challenge~\cite{arc}, HumanEval~\cite{human_eval}, and Grade School Math (GSM8K)~\cite{gsm8k}.

\paragraph{Approximate Retraining.} To attain ideal unlearning outcomes, one can retrain from scratch to exclude the specified sequences $\mathcal{U}$. 
This exact unlearning approach is considered as the ``gold standard'' for evaluating approximate unlearning efficacy, as highlighted in the prior work~\cite{corr/LiuZ23}.
However, retraining LLMs on the entire retained dataset $\mathcal{D} \setminus \mathcal{U}$ is impractical due to substantial computational resource requirements.

To circumvent this, we introduce a surrogate evaluation approach called \emph{approximate retraining}, which is inspired by membership inference attacks~\cite{ShokriSSS17, sp/CarliniCN0TT22} that identify performance gaps between the training and unseen data. We thus hypothesize that the retrained model will exhibit consistent performance on unseen domain-specific data, albeit inferior to its performance on trained data.
Given the significant imbalance between pre-training and unlearning data volumes, we expect the retrained model's performance on unseen data within unlearned data distributions to closely align with the original (vanilla) model's performance.
Consequently, by collecting new data from the same domain as the forget set to create an "approximate set," we estimate the retrained model's performance on the forget set by the vanilla model's performance on this approximate set. 
This estimation can further guide the extent of approximate unlearning, including factors such as the learning rate or optimization steps. 
This approximate set can be easily obtained from the test set of the LLM for its developer or from the open-source community\footnote{\url{https://huggingface.co/RealTimeData}} for all researchers across several domains.

\paragraph{Membership Inference Attack.} The complexity of LLMs precludes straightforward interpretable verification of the complete exclusion of specific sequences from the vanilla model. To address this, we employ Membership Inference Attack (MIA) to ascertain whether particular sequences are erased from the LLM’s training dataset. 
This evaluation employs the Min-K\% Prob method~\cite{shi2023detecting}, which operates on the premise that non-member examples are more prone to containing outlier words with notably high negative log-likelihood values, in contrast to member examples.
An important variable affecting the efficacy of MIA is the percentage of tokens with minimal prediction probability; thus, we conduct experiments across various percentages, selecting the one yielding the highest detection performance for each model. 
The sequence length is set to be 4096 tokens for both member (chunked from the forget set) and non-member (equivalent number of chunks from the approximate set) datasets. 
The effectiveness of MIA was quantitatively assessed using the Area Under Curve (AUC) metric. Notably, a higher AUC indicates that the targeted sequence is still identifiable within the training set, whereas a score approaching 0.5, indicative of random guess result, suggests superior unlearning effectiveness.

\begin{table}[!t]
    \centering
    \small 
    \begin{tabular}{lcccc}
        \toprule
        \multirow{2}{*}{Domains} & \multicolumn{2}{c}{Forget} & \multicolumn{2}{c}{Approximate} \\
        \cmidrule(lr){2-3} \cmidrule(lr){4-5}
        & Docs & Chunks & Docs & Chunks \\
        \midrule
        arXiv & 500 & 1,938 & 6,155 & 32,144 \\
        GitHub & 2,000 & 2,730 & 15,815 & 18,929 \\
        Books & 100 & 3,038 & 50 & 923 \\
        \bottomrule
    \end{tabular}
    \caption{Document and chunk counts across domains.}
    \label{tab:dataset_summary}
\end{table}

\subsection{Model and Datasets}
 We conduct experiments using the open-sourced Yi-6B~\cite{young2024yi} LLM. 
 To rigorously assess the effectiveness of unlearning methods, we perform tests in three distinct settings: arXiv papers, GitHub code repositories, and books. Despite their public availability, these sources may still entail copyright concerns.
 For instance, although arXiv provides open access to preprints, in many cases, the copyright of each individual preprint remains with its authors or the rights holders. 
 Imagine a case where the paper's authors would like to erase their preprints from the pre-trained LLMs (``the right to be forgotten'').

The \textit{forget set} is randomly sampled from the Yi-6B's pre-training data~\footnote{We contacted companies with open-sourced LLMs for pre-training data access. Only the Yi model's developers responded, granting us sampled data and permission for its use and open-sourcing.}, encompassing domains such as arXiv papers, GitHub code repositories, and books. 
Due to the impracticality of evaluating the model's performance across the entirety of the retained set, we randomly select a sample of 1k sequences from the retained set to create a \textit{general set}. 
For arXiv papers, the approximate data comprises 6.1k publications from August 2023. 
The GitHub code repositories' approximate data are 15.8k files from GitHub repositories uploaded in November 2023 with permissive licenses~\cite{realtimedata_github_latest_2024}. 
The approximate data of Books are 50 books published after 2023, which are from the unseen data of BookMIA~\cite{shi2023detecting}.
We employ the model's maximum input sequence length of 4096 as the chunk length, segmenting the sequences into multiple chunks. 
All the approximate dataset is preprocessed in the same manner as~\citet{together2023redpajama}, an open-source pre-training data collection to reproduce Llama. 
The dataset statistics are shown in Table~\ref{tab:dataset_summary}.

\begin{figure*}[!t]
  \centering
  \includegraphics[width=1\linewidth]{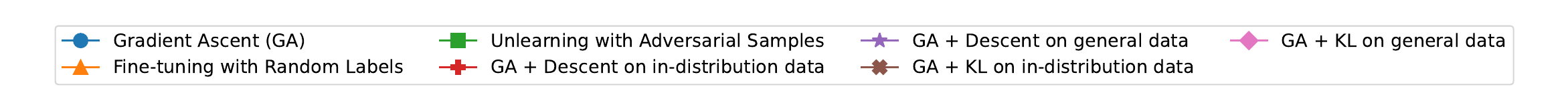}
  % \vspace{-20mm}
  \subfloat[Forget Set PPL]{%
    \includegraphics[width=0.32\linewidth]{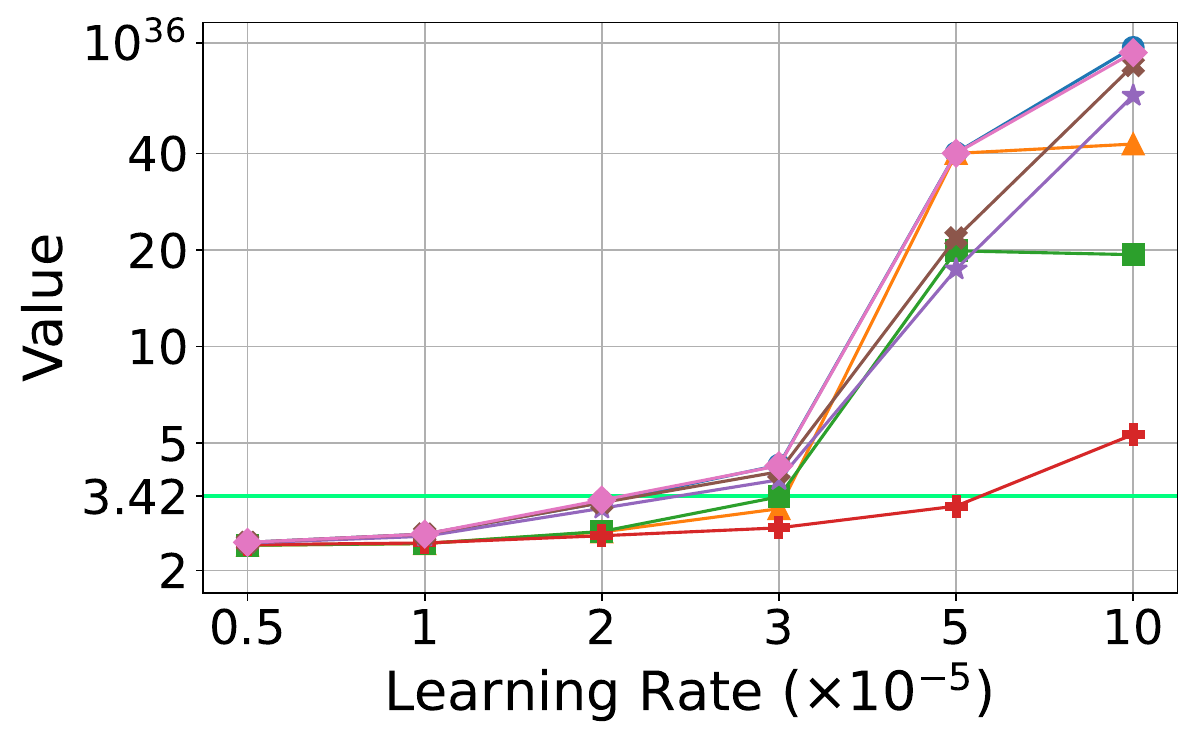}
    \label{fig:1a}
  }\hfill 
  \subfloat[Sampled General Set PPL]{%
    \includegraphics[width=0.32\linewidth]{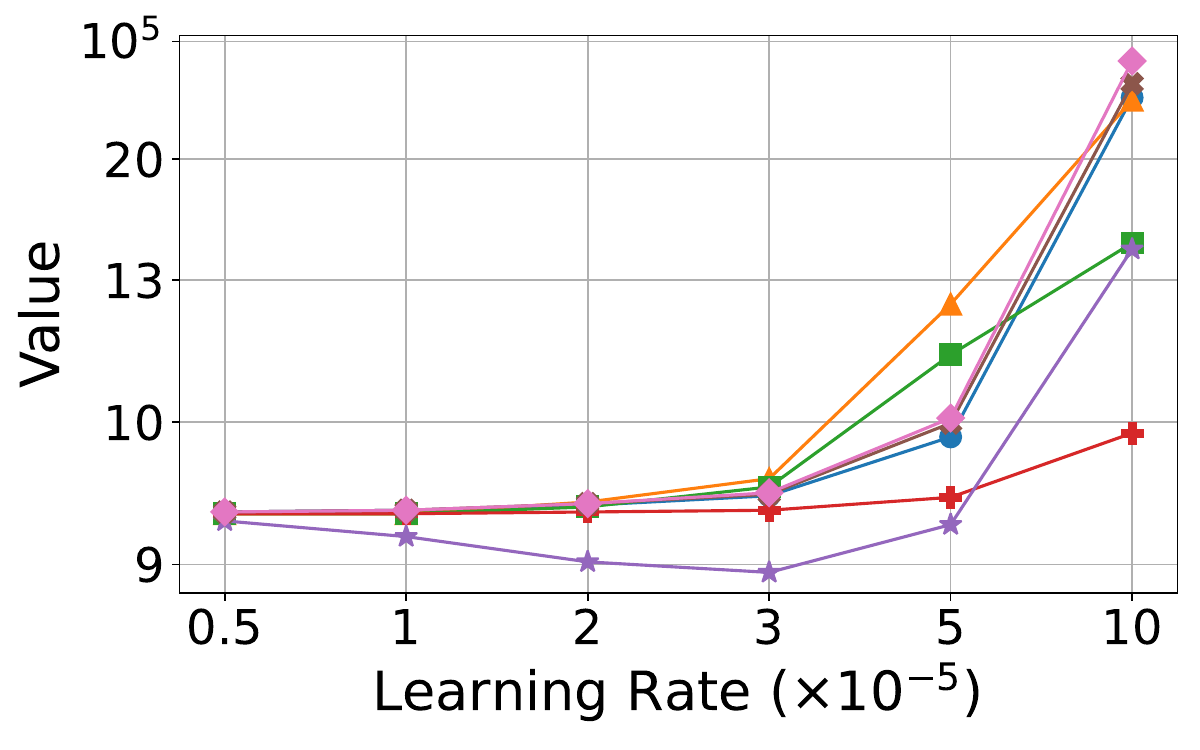}
    \label{fig:1b}
  }\hfill
  \subfloat[Avg. Downstream Tasks performance]{%
    \includegraphics[width=0.32\linewidth]{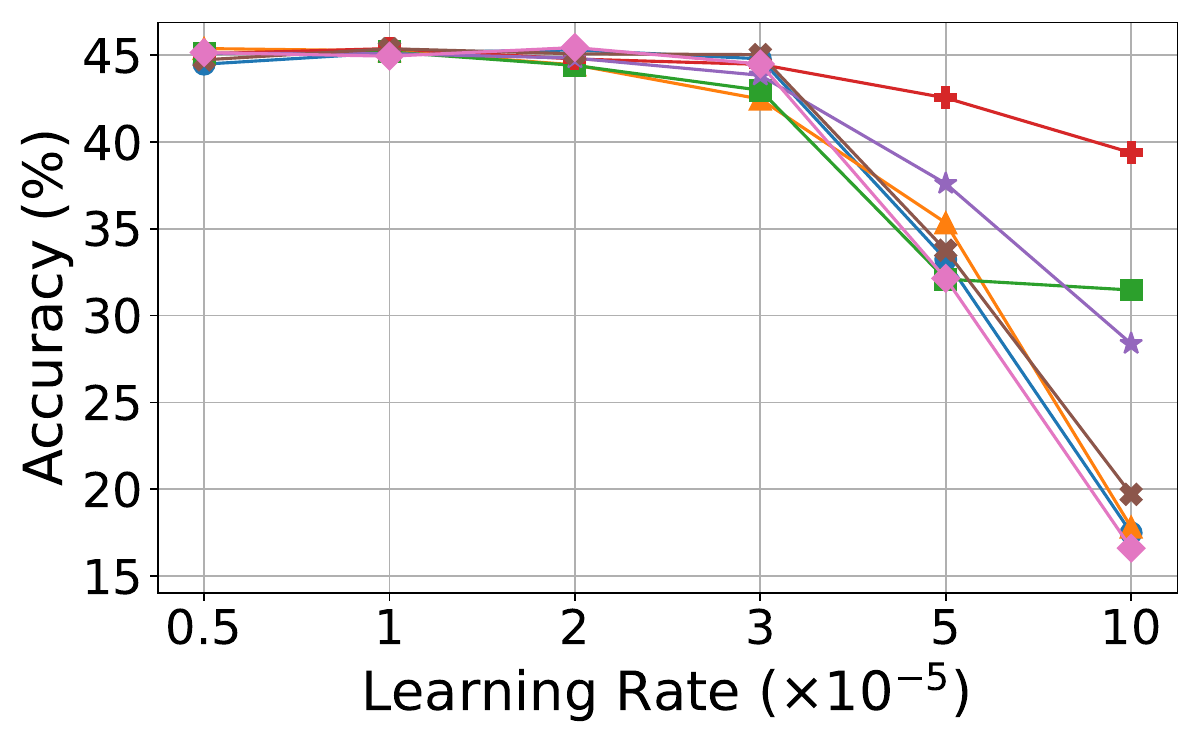}
    \label{fig:1c}
  }

\vspace{1pt}
  
    \subfloat[Forget Set PPL]{%
    \includegraphics[width=0.32\linewidth]{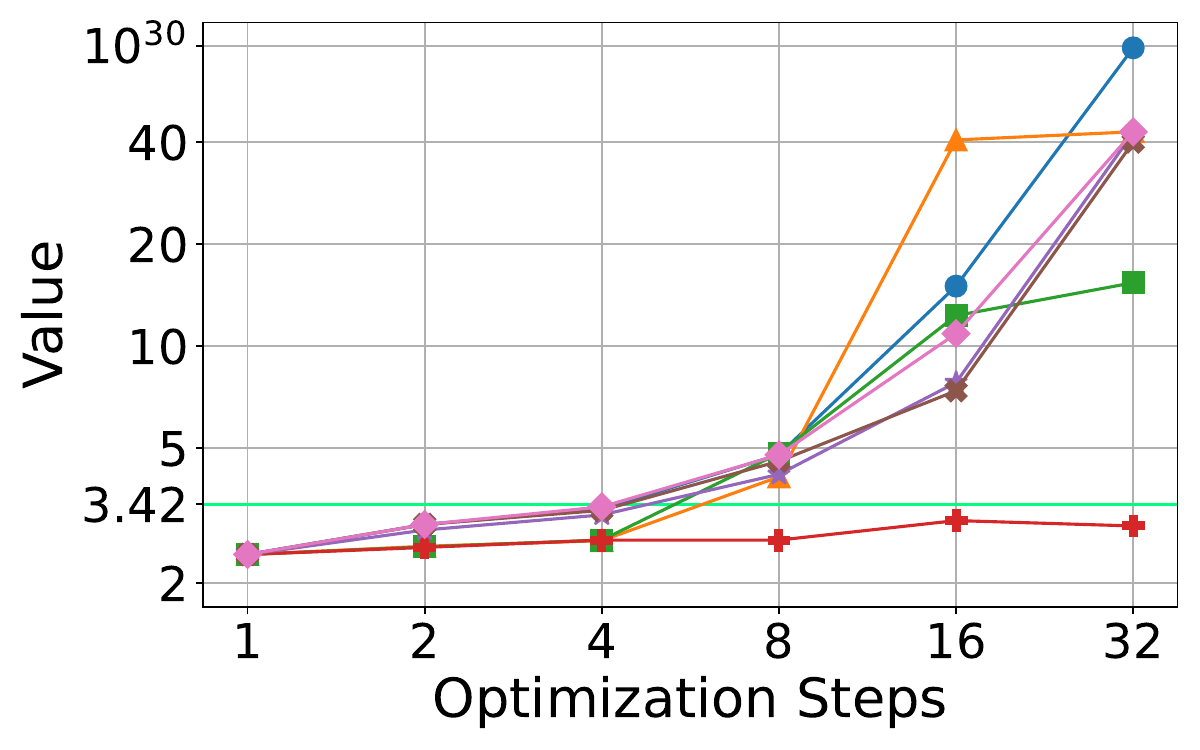}
    \label{fig:1d}
  }\hfill 
  \subfloat[Sampled General Set PPL]{%
    \includegraphics[width=0.32\linewidth]{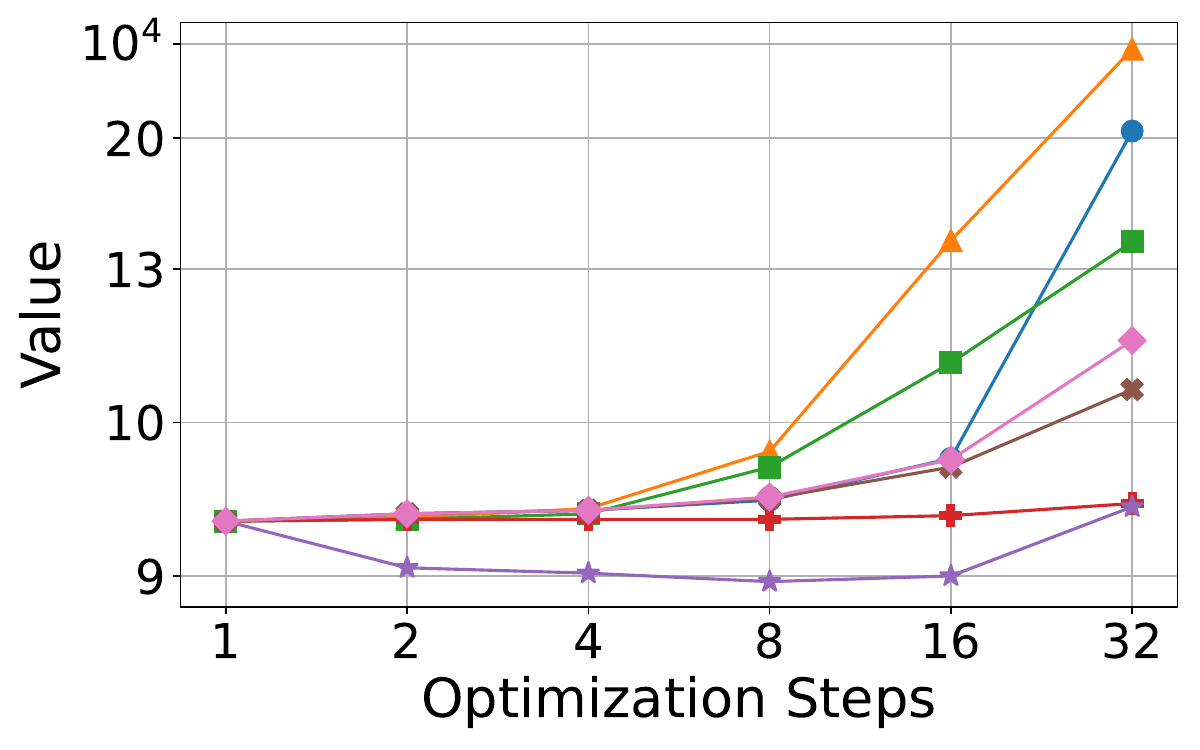}
    \label{fig:1e}
  }\hfill
  \subfloat[Avg. Downstream Tasks performance]{%
    \includegraphics[width=0.32\linewidth]{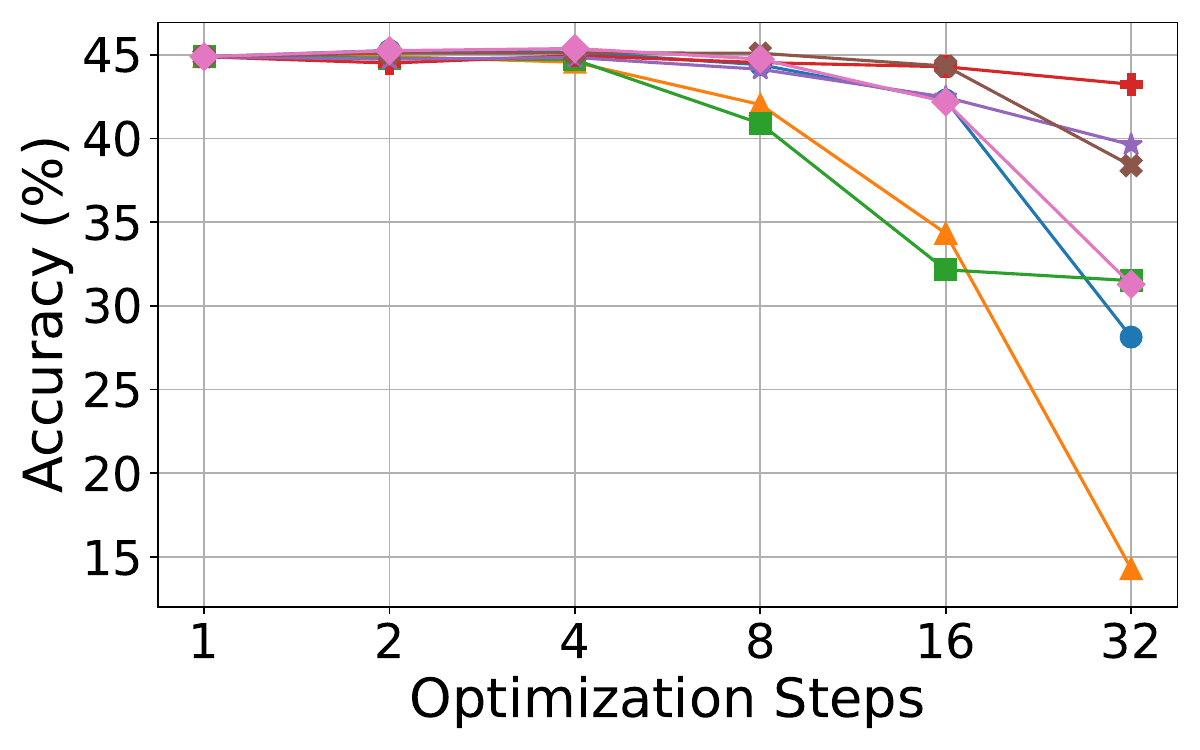}
    \label{fig:1f}}
  \caption{~\cref{fig:1a,fig:1b,fig:1c} are visualization of unlearning results on GitHub code across varying learning rates while optimization steps are fixed at 4. ~\cref{fig:1d,fig:1e,fig:1f} are visualization of unlearning results on GitHub code across varying optimization steps while learning rate is fixed at $2\times 10^{-5}$. In~\cref{fig:1a,fig:1d}, values below 40 are presented on a $\log_{10}$ scale, whereas values above 40 adopt a $\log_{10^{100}}$ scale. The horizontal spring-green line in~\cref{fig:1a,fig:1d} delineates the approximate retraining baseline as the unlearning target. For~\cref{fig:1b,fig:1e}, the scale transitions from $\log_{10}$ for values under 20 to $\log_{10^{10}}$ for values exceeding 20.}
  \label{fig:hyperparameter}
\end{figure*}

\subsection{Results}
We report and analyze the results for unlearning arXiv papers, GitHub code repositories, and books. 
Given that approximate retraining serves as an optimal target for unlearning, we adjust the learning rate of each experiment to align the results with those achieved through approximate retraining. 
The number of unlearning epochs is 1. 
% Considering that most concurrent LLMs pre-training for only one epoch on the dataset~\cite{corr/TouvronMSA+23}, we accordingly limit the unlearning process to a single epoch for all unlearning experiments. 
All the experiments are conducted using 8 A800~GPUs. 

\paragraph{Unlearning academic papers from arXiv.}
We task the pre-trained Yi-6B to unlearn 500 academic papers randomly selected from its training data within the arXiv domain. This procedure simulates scenarios in which authors wish to safeguard their proprietary knowledge or unique writing styles. The unlearning performance is shown in Table~\ref{tab:unlearn_train_data_results_arxiv}.

The vanilla model exhibits close performance on both the forget set and unseen approximate data, with perplexity values of 3.65 and 3.69, respectively, demonstrating the model has good generalization capabilities within the arXiv domain. Compared with the vanilla model, unlearned models exhibit a slight decline in next token prediction accuracy (\eg, from 69.02\% to 68.79\% after gradient ascent), signifying increased difficulty in token extraction given preceding tokens as prompts. 
Notably, only the model subjected to unlearning through a combination of gradient ascent and gradient descent exhibited reduced perplexity on the retain set. This outcome can be attributed to the model learning on a general dataset sampled from the same distribution as the retain set while unlearning the forget set. Furthermore, gradient ascent emerges as the sole method to enhance average accuracy across downstream tasks post-unlearning, underscoring its superiority in maintaining the model's overall utility.

Both the MIA AUC scores before and after unlearning are close to 0.5. This result aligns with a recent study~\cite{duan2024membership}, which indicates that current MIA methods struggle to differentiate between member and non-member data in LLMs. Despite this general difficulty, our results demonstrate a decrease in MIA AUC scores after unlearning.  This indicates that it becomes more challenging to differentiate between the forget set and unseen data, suggesting that our unlearning methods are effective in reducing privacy leakage related to membership inference.

\paragraph{Unlearning programming code from GitHub repositories.}
We request the pre-trained Yi-6B to unlearn 2000 GitHub code files randomly selected from its training data. 
This simulates scenarios where developers or organizations seek to remove specific coding patterns, algorithms, or proprietary code from the model's knowledge base, ensuring that their intellectual property remains protected. The unlearning results are displayed in Table~\ref{tab:unlearn_train_data_results_github}.

The vanilla model exhibits a more pronounced disparity in performance between the forget set and the unseen approximate set for GitHub code, indicating inferior generalization capabilities compared to the arXiv domain. 
A notable decline in the HumanEval pass@1 score is observed for models unlearned through fine-tuning with random labels, unlearning with adversarial samples, and gradient ascent combined with gradient descent on in-distribution data. Given that HumanEval is a metric specifically designed to assess the code-generating proficiency of LLMs~\cite{human_eval}, this substantial decrease underscores the detrimental impact of these three methods on the model's task-specific utility. Furthermore, a performance reduction of at least 1.83\% on the HumanEval pass@1 score is recorded for other methods, suggesting that unlearning GitHub codes from the LLM's pre-training dataset while maintaining coding capabilities presents a challenging task.

\paragraph{Unlearning copyrighted books.}
We instruct the pre-trained Yi-6B to unlearn 500 books randomly selected from its training data. 
This process simulates scenarios in which authors or publishers aim to withdraw their literary works, thereby protecting the uniqueness of their content or preventing the model from generating derivative works. 
The unlearning results are detailed in Table~\ref{tab:unlearn_train_data_results_books}.

% Similarly with the GitHub code domain, the vanilla model exhibits perplexities of 7.62 and 10.11 for the forget set and approximate set, respectively, indicating lesser generalization capabilities compared to the arXiv domain.  Furthermore, when adjusting the unlearning process to achieve perplexity levels akin to those observed in approximate retraining, models unlearned through fine-tuning with random Labels, unlearning with adversarial samples, and gradient ascent combined with gradient descent on in-distribution data demonstrate a lower accuracy on forget set than other methods. This observation implies that these strategies more effectively obfuscate the model's ability to predict subsequent tokens accurately within the forget set, thereby reducing the risk of information extraction from the model.
Similar to the GitHub code domain, the vanilla model's perplexities of 7.62 for the forget set and 10.11 for the approximate set suggest weaker generalization compared to the arXiv domain. Moreover, when fine-tuning to reach perplexity levels similar to approximate retraining, models unlearned via fine-tuning with random labels, adversarial sample unlearning, and the combination of gradient ascent with gradient descent on in-distribution data result in lower accuracy on the forget set than other methods. This indicates that these strategies more effectively obfuscate the model to predict accurately for the forget set, thus better protecting against potential information extraction from the model.

\subsection{Computational efficiency analysis}
Machine unlearning is to make the model efficiently forget specific training data without the need for retraining from scratch. Therefore, efficiency is a crucial factor in evaluating unlearning methods.  Following~\citet{brown2020language}, we estimate the total training floating point operations (FLOPs) as $6 \times \text{Total Training Tokens} \times \text{Parameter Size}$, and the total forward FLOPs as $2 \times \text{Total Forward Tokens} \times \text{Parameter Size}$. For the Yi-6B model, pre-trained on 3 trillion tokens, we calculate the computational costs for retraining and all the unlearning methods applied to forget a dataset comprising 2,000 sequences, each 4,096 tokens long. The results, shown in Table~\ref{tab:flops}, indicate that unlearning is approximately $10^5$ times more efficient than retraining in terms of computational resources.

\begin{table}[!t]
    \centering
    \small 
    \begin{tabular}{l|l}
    \toprule
        \textbf{Methods} & \textbf{FLOPs} \\ \midrule
        Retraining & $1.08\times10^{23}$ \\ 
        Gradient Ascent (GA) & $2.95\times10^{17}$   \\ 
        Fine-tuning with Random Label & $2.95\times10^{17}$ \\ 
        Unlearning with adversarial samples & $3.93\times10^{17}$ \\ 
        GA + Descent on in-distribution data & $5.90\times10^{17}$ \\ 
        GA + Descent on general data & $5.90\times10^{17}$ \\ 
        GA + KL on in-distribution data  & $5.90\times10^{17}$ \\ 
        GA + KL on general data  & $5.90\times10^{17}$ \\ \bottomrule
    \end{tabular}
    \caption{Computational costs across methods. }
    \label{tab:flops}
\end{table}
\subsection{Ablation studies}
\label{subsec: ablation}
Taking the task of unlearning GitHub code as a case study, a series of experiments are conducted to investigate the influence of learning rate and optimization steps on the unlearning outcomes, with results shown in Figure~\ref{fig:hyperparameter}. For~\cref{fig:1a,fig:1b,fig:1c}, we keep optimization steps fixed at 4 and vary the learning rate between $5 \times 10^{-6}$ and $1 \times 10^{-4}$. In~\cref{fig:1d,fig:1e,fig:1f}, the learning rate is constant at $2 \times 10^{-5}$, with optimization steps ranging from 1 to 32.
We evaluated the unlearned model's perplexity on both the forget and general sets, along with average performance on downstream tasks, presenting results across seven different unlearning methods for each hyperparameter configuration. The discussion of the effect of batch size and forget set size on the unlearning results is deferred to Appendix~\ref{app:ablation}.

Figure~\ref{fig:hyperparameter} shows that the method combining gradient ascent and descent on in-distribution data is notably tolerant to changes in learning rate and number of optimization steps, indicating high stability. In contrast, other methods exhibit a marked increase in perplexity for both the forget and general sets with higher learning rates or more optimization steps, underscoring their sensitivity to hyperparameter adjustments.

Moreover, since approximate unlearning may either be non-convergent or compromise utility until convergence, it is crucial to conduct a thorough search for the appropriate hyperparameters to ensure optimal unlearning performance. However, the vast search space and lack of definitive reference targets render this task impractical.
To address these challenges, we analyze and summarize the following detailed \emph{guidelines} to streamline the hyperparameter adjustment process:

% Based on figure~\ref{fig:1d}, a large number of optimization steps makes the unlearning process less stable. However, a small number of optimization steps may lead to the averaging of detailed information across large batch sizes, thus lowering the unlearning quality. Consequently, a balanced optimization step size is proposed to be four.
Figure~\ref{fig:1d} indicates that a high number of optimization steps reduces the stability of the unlearning process, whereas too few steps can average out detailed information over large batches, thereby degrading unlearning quality. Based on these observations, we set the optimization step size to four for optimal balance.

Figure~\ref{fig:1a} shows that the unlearned model's perplexity on the forget set rises with the learning rate. To find the optimal learning rate aligned with the approximate retraining baseline for unlearning, we recommend starting with a broad granularity search ($10^{-5}$) within $5 \times 10^{-6}$ to $5 \times 10^{-5}$. This step narrows the search range. A subsequent finer granularity search within this refined interval will identify the learning rate that best achieves the desired unlearning outcomes.

%% file: 5-related.tex
\section{Related Work}
We provide an overview of current research on machine unlearning, memorization, and forgetting. 
A more detailed version is deferred to Appendix~\ref{app:related_work}.

\paragraph{Machine Unlearning.} The concept of machine unlearning is first introduced in~\citet{sp/CaoY15}. 
% They transform learning algorithms into a summation form and then remove the transformations of the targeted forget sample from these summations. 
\citet{sp/BourtouleCCJTZL21} further formalizes \emph{exact} unlearning by introducing a general framework: sharded, isolated, sliced, aggregated (SISA). Exact unlearning requires the unlearned model the same as the retrained model. \emph{Approximate} unlearning, which relaxes the requirement, is also explored by bounding the distance~\cite{icml/ChourasiaS23} or indistinguishability~\cite{nips/SekhariAKS21} between the two model's distributions. 

Machine unlearning has been extensively researched within the broader field of machine learning~\cite{csur/XuZZZY24}, yet its exploration in generative language models remains limited. 
\citet{corr/KumarGR22} propose SISA-FC and SISA-A, two computationally efficient extensions of SISA for classification~LMs, \eg, BERT.
% \citet{corr/ChenY23} propose an efficient unlearning method via a selective teacher-student formulation for both classification and summarization tasks. 
% \citet{corr/ChenY23} propose an efficient unlearning method for both classification and summarization tasks.
To unlearn knowledge in generative models, 
\citet{acl/JangYYCLLS23} simply perform gradient ascent on target sequences. 
% \citet{wang2024selective} presents a selective unlearning method to minimize negative impacts on unlearned model's capabilities.
\citet{corr/EldanR23} consider a special case of unlearning the Harry Potter books from Llama2-7b.
\citet{yao2023large} applies machine unlearning for harmful responses removing and hallucinations eliminating. However, these studies have been limited to fine-tuned models or a single corpus source. Our work explores unlearning pre-trained LLMs on more diverse datasets. 

\paragraph{Memorization and Forgetting.} 
\citet{uss/Carlini0EKS19} first quantifies \emph{unintended} memorization by a metric called exposure, revealing severe privacy issues, \eg, membership inference attacks~\cite{sp/CarliniCN0TT22} or verbatim data extraction~\cite{uss/CarliniTWJHLRBS21}.
Contrary to memorization, catastrophic forgetting, where a model loses previously learned knowledge when training on new data, has been studied~\cite{aaai/KemkerMAHK18,acl/Shao022}. 
It is a \emph{passive} phenomenon different from unlearning, which actively forces models to ``forget'' specific samples.

%% file: 6-conclusion.tex
\section{Conclusion}
In this paper, we investigate the challenge of removing copyrighted pre-training data from LLMs. We present a unified formulation for unlearning LLMs, from which seven unlearning methodologies are derived. We introduce approximate retraining as an evaluation technique to bypass the impracticality of retraining LLMs from scratch. Our experimental analysis across three pre-training data domains validates the efficacy of the unlearning approaches. Furthermore, we find that combining gradient ascent with gradient descent on in-distribution data enhances hyperparameter robustness. We also offer guidelines to streamline the tuning of hyperparameters essential to the unlearning process.

%% file: app-ablation.tex
\section{Additional Ablation Studies}
\label{app:ablation}

\begin{figure*}[!t]
  \centering
  \includegraphics[width=1\linewidth]{figures/legend.pdf}
  % \vspace{-20mm}
  \subfloat[Forget Set PPL]{%
    \includegraphics[width=0.32\linewidth]{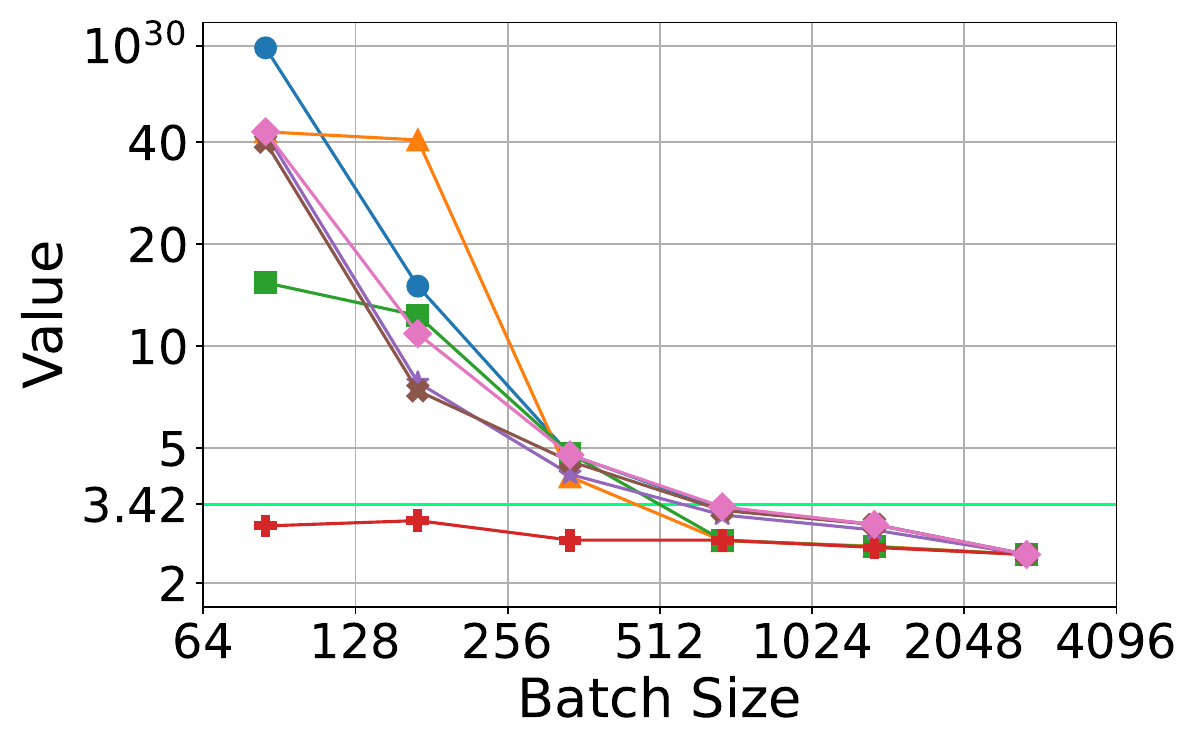}
    \label{fig:2a}
  }\hfill 
  \subfloat[Sampled General Set PPL]{%
    \includegraphics[width=0.32\linewidth]{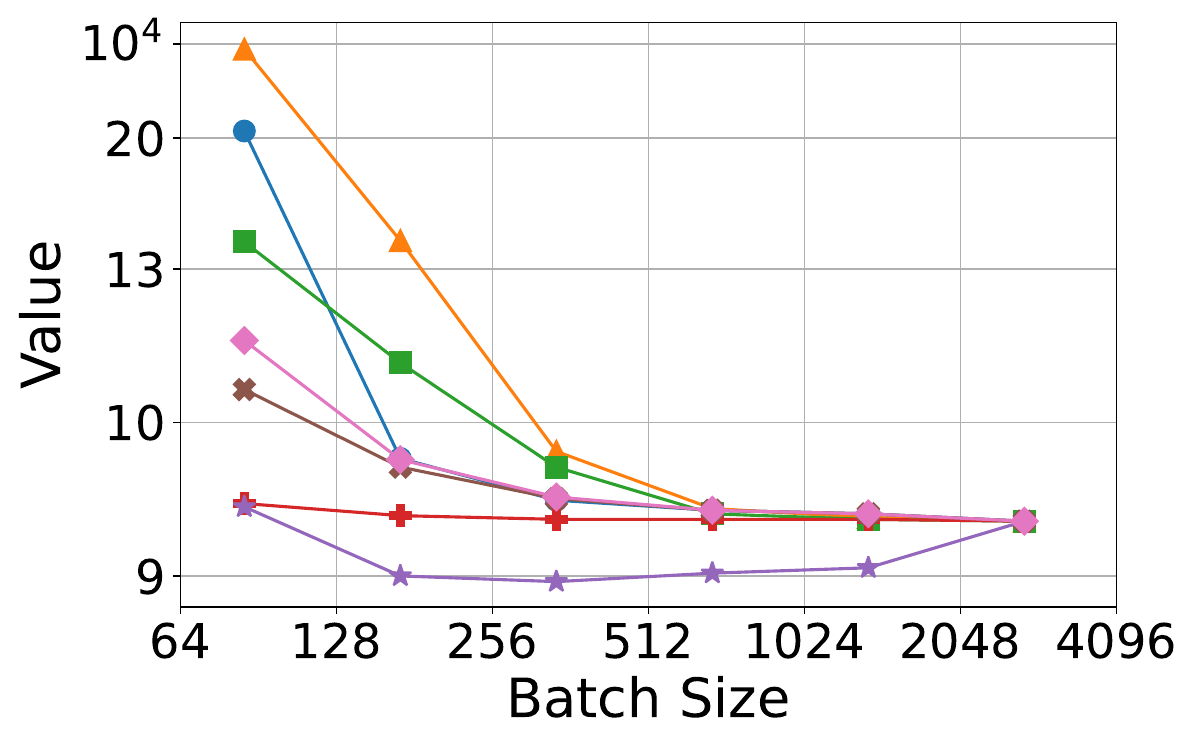}
    \label{fig:2b}
  }\hfill
  \subfloat[Avg. Downstream Tasks performance]{%
    \includegraphics[width=0.32\linewidth]{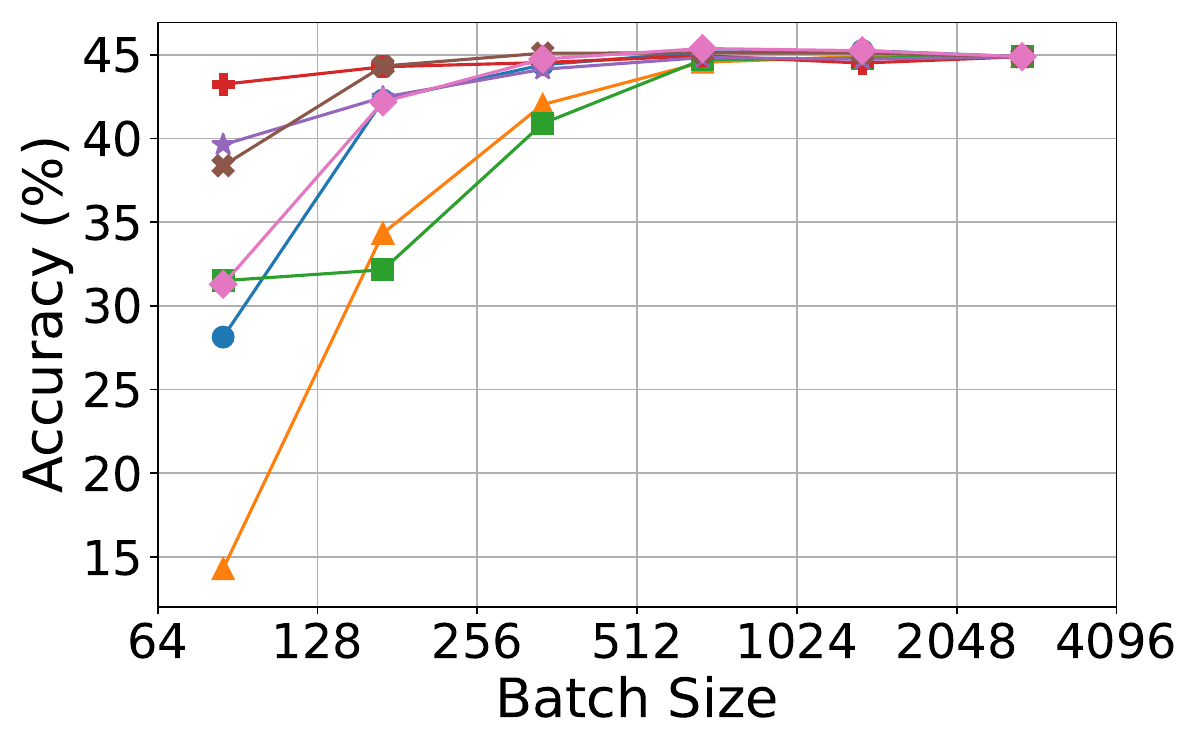}
    \label{fig:2c}
  }

\vspace{1pt}
  
    \subfloat[Forget Set PPL]{%
    \includegraphics[width=0.32\linewidth]{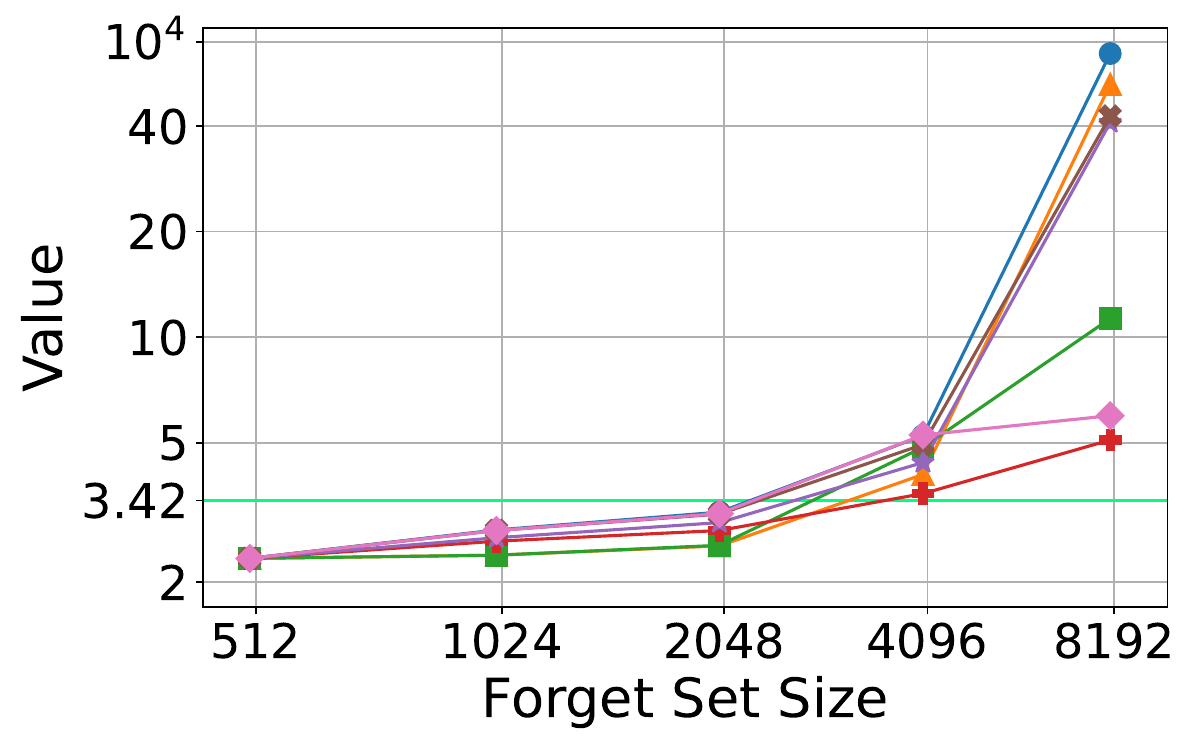}
    \label{fig:2d}
  }\hfill 
  \subfloat[Sampled General Set PPL]{%
    \includegraphics[width=0.32\linewidth]{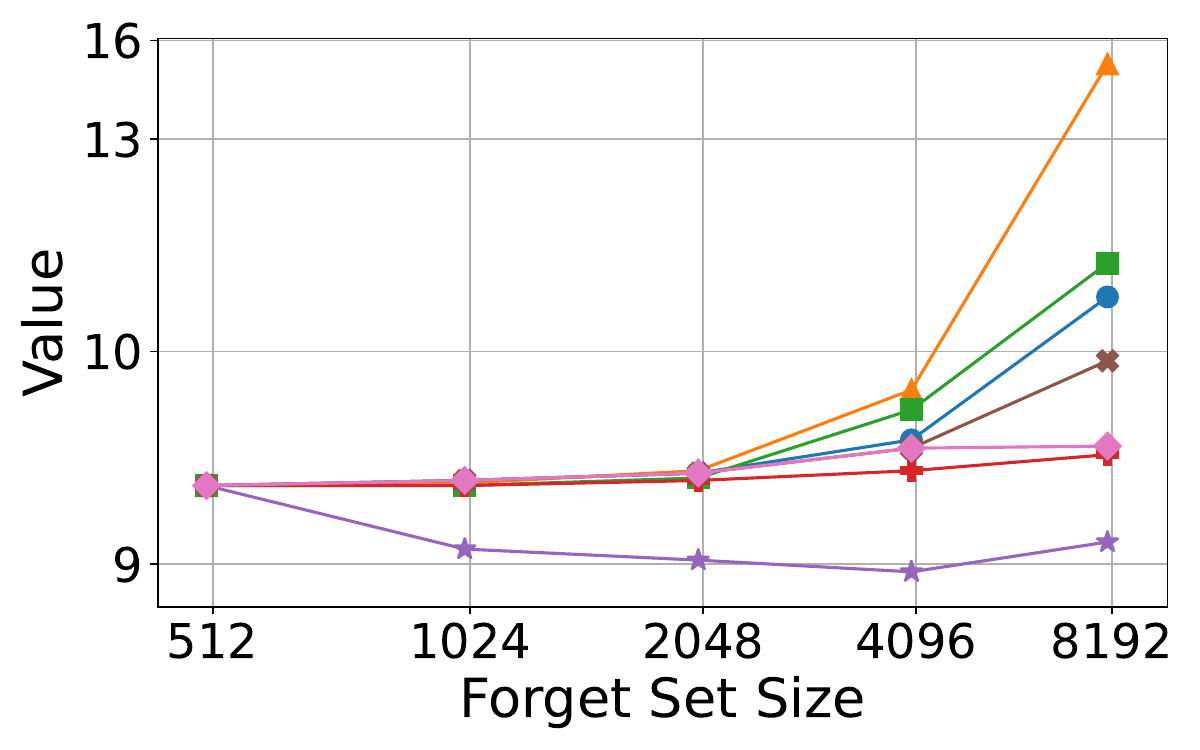}
    \label{fig:2e}
  }\hfill
  \subfloat[Avg. Downstream Tasks performance]{%
    \includegraphics[width=0.32\linewidth]{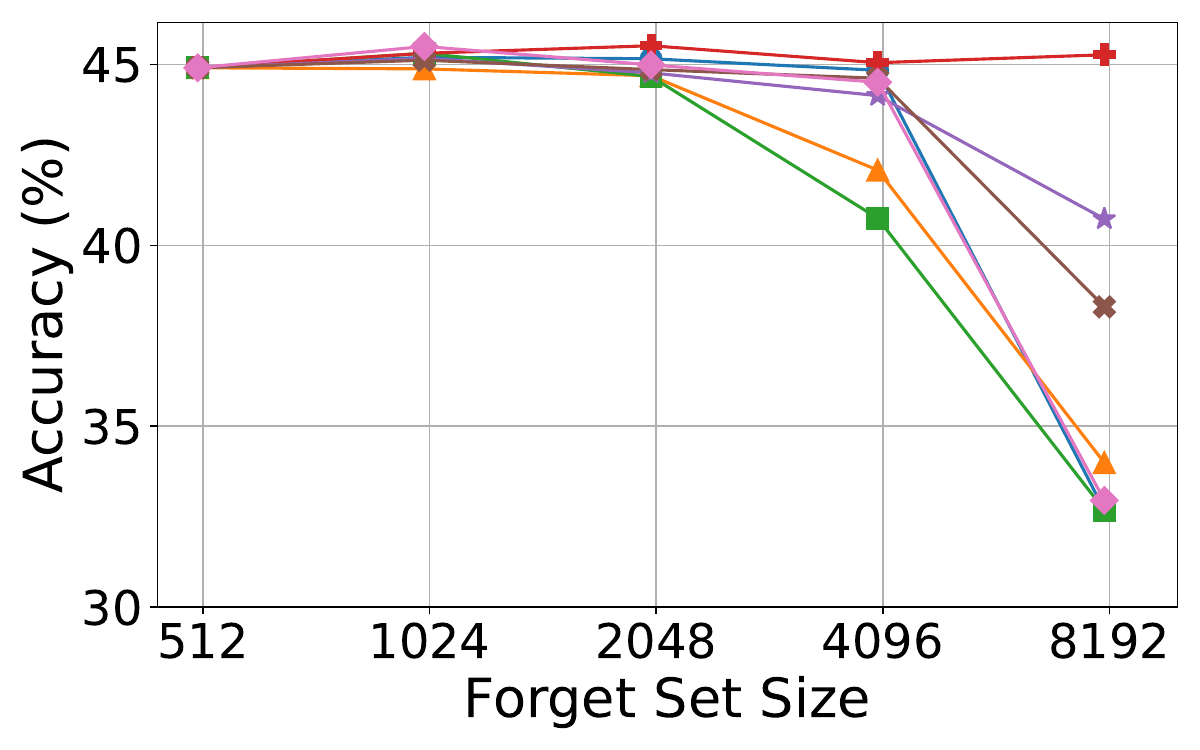}
    \label{fig:2f}}
  \caption{~\cref{fig:2a,fig:2b,fig:2c} are visualization of unlearning results on GitHub code across varying batch sizes while other hyperparameters are fixed. ~\cref{fig:2d,fig:2e,fig:2f} are visualization of unlearning results on GitHub code across varying forget set size while other hyperparameters are fixed. In~\cref{fig:2a,fig:2d}, values below 40 are presented on a $\log_{10}$ scale.  Values in \cref{fig:2a} above 40 adopt a $\log_{10^{100}}$ scale, whereas values in \cref{fig:2d} above 40 adopt a $\log_{10^{10}}$ scale.
  The horizontal spring-green line in~\cref{fig:2a,fig:2d} delineates the approximate retraining baseline as the unlearning target. For~\cref{fig:2b,fig:2e}, the scale transitions from $\log_{10}$ for values under 20 to $\log_{10^{10}}$ for values exceeding 20.}
  \label{fig:app-ablation}
\end{figure*}

~\cref{fig:2a,fig:2b,fig:2c} display the results of unlearning 2720 sequences of GitHub code, with the batch size varied from 85 to 2720 and other hyperparameters fixed. The results demonstrate that a small batch size leads to more optimization steps, which can cause detrimental damage to the model’s general utility. However, a higher batch size also reduces the unlearning efficacy. As mentioned in Section~\ref{subsec: ablation}, a batch size with corresponding optimization steps of four would be the best practice to balance utility and unlearning effect.

~\cref{fig:2d,fig:2e,fig:2f} show the results of unlearning GitHub code with varying forget set sizes (512 to 8192 sequences) and fixed hyperparameters. The results show that excessively large unlearning data sizes lead to an exponential increase in perplexity on both the forget and general sets and also cause a significant decrease in model general utility. Integrating gradient ascent with gradient descent on in-distribution data proves most stable, aligning with insights in Section~\ref{subsec: ablation}. For unlearning large datasets, we recommend using a relatively large batch size for stability.

%% file: app-random_seed.tex
\begin{table*}[!t]
\resizebox{\linewidth}{!}{%
\begin{tabular}{@{}l|ccc|cc|ccccc@{}}
\toprule
\multirow{2}{*}{Models} & \multicolumn{3}{c|}{Forget Set} & \multicolumn{2}{c|}{Retain Set} & \multicolumn{5}{c}{Downstream Task Accuracy $\uparrow$} \\ \cmidrule(l){2-11} 
 & ACC$\downarrow$ & PPL$\uparrow$ & MIA$\downarrow$ & ACC$\uparrow$ & PPL$\downarrow$ & MMLU & ARC & HumanEval &  GSM8K & Avg. \\ \midrule
 Vanilla Model & 80.65 & 2.40 & 81.93 & 52.68 & 9.24 & 63.37 & 68.49 & 16.46 & 33.59 & 45.48 \\
Approximate Retrain & 72.91 & 3.42  & - & - & - & - & - & - & - & -  \\
Gradient Ascent & 78.18 $\pm$ 0.03 & 3.56 $\pm$ 0.03  &74.23 $\pm$ 0.05  & 52.6 $\pm$ 0.01 & 9.31 $\pm$ 0 & 63.47 $\pm$ 0.05 &68.23 $\pm$ 0.14  & 14.83 $\pm$ 0.93 & 34.77 $\pm$ 0.39 & 45.33 $\pm$ 0.16 \\
Fine-tuning with Random Labels &77.98 $\pm$ 0.02  & 3.12 $\pm$ 0.01 &80.52 $\pm$ 0.03  & 52.51 $\pm$ 0.01 & 9.47 $\pm$ 0 & 62.43 $\pm$ 0.09 & 67.13 $\pm$ 0.1 & 11.39 $\pm$ 0.35 & 29.74 $\pm$ 0.94 & 42.67 $\pm$ 0.3  \\
Unlearning with Adversarial Samples & 75.17 $\pm$ 0.08& 3.39 $\pm$ 0.01 & 79.54 $\pm$ 0.13 & 52.54 $\pm$ 0.01 & 9.41 $\pm$ 0 &  62.34 $\pm$ 0.05 & 67.41 $\pm$ 0.13 & 10.78 $\pm$ 0.93 & 31.64 $\pm$ 0.27 &  43.04 $\pm$ 0.3  \\ \midrule
Gradient Ascent + Descent on retain set &  &  &  &  &  &  &  &  &  &  \\
- Descent on in-distribution data & 76.91 $\pm$ 0.03 & 3.44 $\pm$ 0.01 & 76.76 $\pm$ 0.01 & 52.48 $\pm$ 0.01 & 9.38 $\pm$ 0 & 62.34 $\pm$ 0.03 & 66.76 $\pm$ 0.02 & 2.85 $\pm$ 0.7 & 30.61 $\pm$ 0.7 &  40.64 $\pm$ 0.01 \\
- Descent on general data & 78.79 $\pm$ 0 & 3.57 $\pm$ 0.01  &75.61 $\pm$ 0.01  & 53.03 $\pm$ 0.01 & 9 $\pm$ 0  & 63.19 $\pm$ 0.05 & 67.56 $\pm$ 0.06 & 15.04 $\pm$ 0.7 & 33.26 $\pm$ 0.58 & 44.76 $\pm$ 0.04 \\ \midrule

Gradient Ascent + KL divergence &  &  &  &  &  &  &  &  &  &  \\
- KL on in-distribution data &78.78 $\pm$ 0 &3.51 $\pm$ 0.01  &76.19 $\pm$ 0  & 52.61 $\pm$ 0.01 & 9.31 $\pm$ 0 & 63.44 $\pm$ 0.04 & 68.23 $\pm$ 0.02 & 14.63 $\pm$ 0 &34.85 $\pm$ 0.25  & 45.29 $\pm$ 0.06 \\
- KL on general data & 78.68 $\pm$ 0& 3.58 $\pm$ 0 & 75.42 $\pm$ 0 & 52.6 $\pm$ 0 & 9.31 $\pm$ 0  & 63.39 $\pm$ 0.06 & 68.01 $\pm$ 0.06 & 14.43 $\pm$ 0.7 & 35 $\pm$ 0.31 & 45.21 $\pm$ 0.18 \\
\bottomrule
\end{tabular}%
}
\caption{Overall results of unlearning an open-source LLM on a subset of pre-training data (2K GitHub code repository files) under random seed 42, 420, and 4200. The values are displayed in the mean $\pm$ standard deviation format.}
\label{tab:unlearn_train_data_results_github_random_seed}
\end{table*}

%% file: app-related.tex
\section{Related Work (Full Version)}
\label{app:related_work}
%\du{\cite{eldan2023whos}: approximate unlearning definition on \emph{subset} (Type II) of training data (specifying a particular behavior/user/...?) vs. a \emph{particular} example/sentence (Type I)?}

In this section, we provide greater detail about related work on machine unlearning, memorization and forgetting, the relation between machine unlearning, differential privacy, and alignment, exact unlearning, and second-order methods of machine unlearning. 

\paragraph{Machine Unlearning.} 
%It has gained significant attention in recent research due to the increasing need for data privacy and the challenges posed by deep learning models.
\citet{sp/CaoY15} introduce the notion of machine unlearning.
They give a heuristic method, transforming learning algorithms into a summation form for forgetting data lineage.
Their goal is to ensure that~unlearned models exactly match the ones retrained from scratch. 
Subsequently, it is formalized as \emph{exact} unlearning or data deletion~\cite{nips/GinartGVZ19,sp/BourtouleCCJTZL21} of a specific training sample, requiring the distributions of unlearned and retrained models are identical.
\citet{nips/GinartGVZ19} propose two tailored approaches for $k$-means, while~\citet{sp/BourtouleCCJTZL21} propose a general unlearning framework: sharded, isolated, sliced, aggregated (SISA).

Exact unlearning may be too ``strong'' to achieve; it can be ``relaxed'' to \emph{approximate} unlearning~\cite{nips/GinartGVZ19} by bounding the ``distance'' (\eg, R{\'e}nyi divergence~\cite{icml/ChourasiaS23} or  indistinguishability~\cite{nips/SekhariAKS21}) between the two models' distributions.
More generally, one can unlearn a subset of training points~\cite{nips/SekhariAKS21} that can even be adaptively chosen~\cite{nips/GuptaJNRSW21}.

%\cite{icml/GuoGHM20} unlearning requires access to the entire training data (for data deletion), whereas \cite{nips/SekhariAKS21} does not.
%Prior arts~\cite{icml/GuoGHM20} focus on empirical risk minimization and training loss, whereas \cite{nips/SekhariAKS21} considers arbitrary convex loss functions and test loss.

Since the seminal proposal, machine unlearning has been widely studied in ML in general~\cite{csur/XuZZZY24,gong2024trajdeleter} but remains rarely explored in NLP, notably generative LLMs.
~\citet{corr/ZhangFTP+23} discusses the challenges and implications of unlearning and other approaches to realize RTBF in LLMs.
\citet{corr/KumarGR22} propose SISA-FC and SISA-A, two extensions of SISA for classification~LMs, \eg, BERT.
SISA-FC only trains fully connected task layers, and SISA-A resorts to Adapters~\cite{icml/HoulsbyGJMLGAG19} that only update a few plug-in parameters.
\citet{corr/ChenY23} propose an efficient unlearning method via a selective teacher-student formulation for both classification and summarization tasks. They also design a fusion mechanism to merge unlearning layers for sequential data forgetting.
To unlearn knowledge in generative models, \citet{acl/JangYYCLLS23} simply perform gradient ascent on target sequences. \citet{wang2024selective} presents a selective unlearning
method to minimize negative impacts on unlearned model's capabilities and proposes evaluation metrics focusing on sensitive information. 
\citet{corr/EldanR23} consider a special case of unlearning the Harry Potter books from Llama2-7b.
They first use a reinforced model to identify the tokens that are most related to the unlearning target and then replace idiosyncratic terms with generic ones to generate alternative labels for fine-tuning the model.
\citet{yao2023large} applies machine unlearning for harmful responses removing and hallucinations eliminating. \citet{maini2024tofu} presents a benchmark for unlearning fictitious authors on fine-tuned models.

\paragraph{Memorization and Forgetting.} 
Training data memorization to some extent is pivotal for model generalization, but \emph{unintended} memorization, first quantified by a metric called exposure~\cite{uss/Carlini0EKS19}, poses severe privacy issues, \eg, membership inference attacks~\cite{sp/CarliniCN0TT22}, verbatim data extraction~\cite{uss/CarliniTWJHLRBS21}, or property inference attacks~\cite{mao2023secure}.
\citet{iclr/CarliniIJLTZ23} illustrates that memorization relies on the model scale, training data deduplication, and prompting context length.

As opposed to memorization, catastrophic forgetting, which means that a model tends to forget previously learned knowledge when training on new data, has been studied~\cite{aaai/KemkerMAHK18,acl/Shao022}. 
It is a \emph{passive} phenomenon different from unlearning, which actively forces models to ``forget'' specific samples.
As with memorization, concurrent works~\cite{nips/TirumalaMZA22,iclr/Jagielski0TILCW23} define and measure forgetting  as a form of privacy leakage.

% Memorization in foundation models is a byproduct of training on extensively large datasets. During training, models learn to capture statistical regularities in the data, which includes memorizing certain facts, phrases, or sequences. This memorization is pivotal for tasks requiring factual accuracy and context retention over multiple interaction steps. But on the other hand, this memorization leads to significant data privacy, copyright, or other regulatory issues. We need to actively unlearn and forget information stored in the foundation models. Existing work in terms of ``forgetting'' mostly focuses on passive forgetting while little work explores a mechanism for active forgetting.   

\paragraph{Machine Unlearning vs. Differential Privacy.}
DP is a rigorous framework for protecting individual privacy in data analytics by adding calibrated noise to query results~\cite{fttcs/DworkR14}, which has been explored in language models~\cite{yue2021differential,yue2022synthetic,li2022large,  du2023sanitizing, du2023dp}.
Definition-wise, approximate unlearning is reminiscent of DP.
They use the same metric for distributional closeness (\eg, $(\epsilon, \delta)$-indistinguishability~\cite{icml/GuoGHM20}) but with a substantial difference. 
Unlearning~compares two algorithms--unlearning and retraining--on the same dataset, whereas DP compares the same algorithm run on neighboring datasets (differing in an individual's data).
DP is a sufficient (not necessary) condition for unlearning~\cite{icml/GuoGHM20}:
An DP mechanism working on datasets with edit distance $m$ naturally unlearns \emph{any} $m$ samples.
Prior DP-based unlearning designs~\cite{icml/GuoGHM20,nips/SekhariAKS21,aistats/IzzoSCZ21} often assume convex loss functions.  
\citet{nips/SekhariAKS21,corr/HuangC23} bound the ``deletion'' capacity (\ie, how many samples can be unlearned while ensuring desirable loss) better than $m$ in DP.
Also, DP can mitigate the adaptivity of unlearning requests~\cite{nips/GuptaJNRSW21}. \citet{eurocrypt/GargGV20} provide an alternative definitional framework for RTBF from cryptographic primitives. 
% DP is also utilized solely to promote privacy in language models. \citet{li2022large} shows LLMs fine-tuned using differentially private stochastic gradient descent can also have nondescending performance by non-standard hyperparameters. \citet{yue2021differential,du2023sanitizing} apply DP on language models fine-tuning via sanitizing natural text and sentence embeddings. \citet{du2023dp} shows directly perturbs embedding matrices in the forward pass of language models could realize DP guarantee in the inference stage. 

\paragraph{Machine Unlearning vs. Alignment.} Alignment in LLMs, the process of adjusting these models to resonate with human values, is typically accomplished through techniques like supervised fine-tuning (SFT)~\cite{nips/Ouyang0JAWMZASR22,DatabricksBlog2023DollyV2,vicuna2023,corr/KopfKR+23,corr/TouvronMSA+23} and reinforcement learning with human feedback (RLHF)~\cite{nips/ChristianoLBMLA17,nips/StiennonO0ZLVRA20,nips/Ouyang0JAWMZASR22,corr/BaiJNA+23,corr/TouvronLIM+23,corr/TouvronMSA+23}. These methods rely on human-generated demonstrations or rewards and penalty systems. Machine unlearning, on the other hand, uniquely focuses not on promoting correct behavior but discouraging outputs misaligned with human values~\cite{yao2023large}. This method thus offers a complementary approach to standard SFT techniques.

\paragraph{Second-order methods of Machine Unlearning.} The high-level idea of almost all second-order unlearning methods is through Taylor expansion on the gradient at the stationary point. This leads to a Newton-type update, which involves Hessian inverse (or its approximation) computation~\cite{cvpr/GolatkarAS20,corr/PesteAL21}. Apparently, the hessian-related operation is prohibitive to LLM due to its billions if not trillions parameters. Nevertheless, current theoretical approximate unlearning approaches with privacy guarantees are all second-order to the best of our knowledge~\cite{icml/GuoGHM20,nips/SekhariAKS21,iclr/Chien0M23}. We decided to briefly introduce them in the context of classification and discuss the potential way of extending them for LLM.

Here, we denote $M$ for both the LLM and its model parameters with a slight abuse of notation. Assume $M$ is well-trained with respect to the training loss $\mathcal{L}(P_M; \mathcal{D})$ so that it is a stationary point $\nabla \mathcal{L}(P_M; \mathcal{D}) = 0$. Similarly, the retrain model $M_r^{\mathcal{U}}$ is also a stationary point with respect to the loss $\mathcal{L}(P_{M_r^{\mathcal{U}}}; \mathcal{D}\setminus\mathcal{U})$, so that $\nabla \mathcal{L}(P_{M_r^{\mathcal{U}}}; \mathcal{D}\setminus\mathcal{U})=0$. We can apply a first order Tayler expansion on $\nabla \mathcal{L}(P_{M_r^{\mathcal{U}}}; \mathcal{D}\setminus\mathcal{U})$ at $M$, which leads to

{
\small
\begin{align*}
    & \nabla \mathcal{L}(P_M; \mathcal{D}\setminus\mathcal{U}) + \nabla^2 \mathcal{L}(P_M; \mathcal{D}\setminus\mathcal{U}) (M_r^{\mathcal{U}}-M) \approx 0\\
    & \Rightarrow M_r^{\mathcal{U}} \approx M -  (\nabla^2 \mathcal{L}(P_M; \mathcal{D}\setminus\mathcal{U}))^{-1}\nabla \mathcal{L}(P_M; \mathcal{D}\setminus\mathcal{U}).
\end{align*}
}

% \begin{align*}
% & \text{GradL} + \text{HessL} \cdot (M_r^{\mathcal{U}}-M) \approx 0\\
% & \Rightarrow M_r^{\mathcal{U}} \approx M - \text{HessL}^{-1}\text{GradL}.
% \end{align*}
The theoretical unlearning approach will further introduce some privacy noise (similar to the Gaussian mechanism~\cite{csfw/Mironov17}) to obfuscate the potential privacy leakage rigorously~\cite{icml/GuoGHM20,nips/SekhariAKS21,iclr/Chien0M23}, where the noise variance determined by the worst-case error of the Taylor approximation. This analysis can only be done for strongly convex problems with additional smoothness assumptions and is thus not applicable to LLMs. In practice, some researchers still follow this second-order update with a heuristic-based noise addition design, which has demonstrated superior performance on privacy and utility~\cite{cvpr/GolatkarAS20}. The main focus of this direction is to improve the computation complexity of the second-order update, with ideas leveraging the Fisher information matrix~\cite{cvpr/GolatkarAS20} and rank one update by Sherman-Morrison lemma~\cite{corr/PesteAL21}. Nevertheless, the computation complexity of these advanced second-order methods is still too expensive for LLM. One potential direction is to apply these second-order updates only for adaptor~\cite{icml/HoulsbyGJMLGAG19} or LoRA~\cite{iclr/HuSWALWWC22}, which contain much fewer parameters to be modified. It remains open whether it is possible to apply second-order methods for LLM or not at the moment. 

\paragraph{Exact Unlearning.} We introduce exact unlearning methods that correspond to $(\epsilon, \delta)$ notion of unlearning with $\epsilon=\delta=0$. While ensuring $\epsilon=\delta=0$ is desired in some extreme cases, it generally fails to explore the beneficial trade-off between unlearning quality, model utility, and time complexity.

% type I unlearning with $\xi=0$. While ensuring $\xi=0$ is desired in some extreme cases, it generally fails to explore the beneficial trade-off between $\xi$, model utility, and time complexity. 

SISA (Sharded, Isolated, Sliced, and Aggregated) framework~\cite{sp/BourtouleCCJTZL21} is a general approach to achieve exact unlearning for general deep neural networks at the cost of changing the training pipeline significantly. The main idea is to partition the training dataset $\mathcal{D}$ into $K$ disjoint sets $\mathcal{D}_1,\ldots,\mathcal{D}_K$. For each $\mathcal{D}_i$, we train or fine-tune $M$ on it independently, which results in $K$ models $M_i=A(\mathcal{D}_i)$. We use any fixed aggregation strategy for these $K$ models for the final prediction or output. Unlearning in the SISA framework is straightforward. Given an unlearning request $\mathcal{U}$, we retrain all models $M_i$ for $i$ such that $\mathcal{U}\cap \mathcal{D}_i\neq \emptyset$. Apparently, storing $K$ copies of LLMs is memory-expensive and impractical which makes the SISA approach not applicable to the pre-training task. The authors of~\cite{corr/KumarGR22} leverage the SISA approach to the fine-tuning task by only fine-tuning a fully connected layer (FC) or Adapter (A) on top of a freeze public pretrained model $M$. They termed these methods SISA-FC and SISA-A respectively.

The SISA framework is currently the only method providing a theoretical privacy guarantee while applicable to LLMs. However, the efficiency and utility of this approach are greatly affected by the choice of $K$ and dataset dependent. Clearly, when $K=1$ we simply arrive at retraining from scratch, which maximally preserves the utility but exhibits impractical time complexity. Choosing a large $K$ can improve the efficiency but may degrade model utility. It is unclear at the moment how to choose an appropriate $K$. 
% On the other hand, the SISA framework focuses on erasing the effect of some training data and naturally compares itself with retraining. It is only applicable to type I but not type II unlearning.

% \textbf{Exact Unlearning for last $k$ layer~\cite{corr/KurmanjiTT23}:} As indicated by its name, the authors propose to freeze all but the last $k$ layers of $M$ and perform retraining. The intuition is that the most relevant information is contained in the last few layers. However, this approach provides neither any formal privacy guarantee nor a comparison with retraining. }